\documentclass{article}

\usepackage[T1]{fontenc}
\usepackage{microtype}
\usepackage{graphicx}
\usepackage{subfigure}
\usepackage{booktabs} %
\usepackage{graphicx}
\usepackage{tikz}
\usepackage{algorithm}
\usepackage{algpseudocode}
\usepackage{amsmath}
\usepackage{amssymb}
\usetikzlibrary{shapes, arrows, positioning, fit, calc, backgrounds}
\usepackage{subcaption}
\usepackage{pgfplots}
\usepackage{listings}
\usepackage{hyperref}

\makeatletter\@namedef{ver@algorithmic.sty}{}\makeatother

\usepackage[accepted]{mlsys2025}
\makeatletter
\renewcommand{\Notice@String}{Preprint (v3), October 30, 2026.}
\makeatother
\hypersetup{pdftitle={Mixture-of-Models: Distributed Inference over Heterogeneous LLMs as Model-Level Mixture-of-Experts}, pdfauthor={Tims Pecerskis, Aivars Smirnovs}, pdfsubject={Preprint}}

\mlsystitlerunning{Mixture-of-Models: Distributed Inference over Heterogeneous LLMs as Model-Level Mixture-of-Experts}

\begin{document}

\twocolumn[
	\mlsystitle{Mixture-of-Models: Distributed Inference over Heterogeneous LLMs as Model-Level Mixture-of-Experts}

	\mlsyssetsymbol{equal}{*}

	\begin{mlsysauthorlist}
        \mlsysauthor{Tims Pecerskis}{to}
        \mlsysauthor{Aivars Smirnovs}{to}
  \end{mlsysauthorlist}

	\mlsysaffiliation{to}{Peeramid Labs The AI Futures Collective}

	\mlsyscorrespondingauthor{Tims Pecerskis}{tim@peeramid.xyz}

	\mlsyskeywords{Machine Learning, MLSys}

	\vskip 0.3in

	\begin{abstract}
		Switch Transformers scale MoE model capacity by routing tokens in memory for one jointly trained model, requiring high-bandwidth interconnects, large memory, centralised ownership, and lack scalable composability. We propose Mixture-of-Models (MoM) architecture as Switch Transformer scaling to semantic level which composes inference across independently trained distributed LLMs, each run on its own GPU. The models reach joint consensus using N-Way Self-Evaluating Deliberation protocol: a combination of social-choice methods such as Delphi method and Quadratic Voting as well as Inference-Time scaling paradigm. We start with an empirical model that characterizes consensus as an inference-time trade-off between signal extraction and contextual noise accumulation, establishing a mathematical basis for convergence driven inference stopping strategies. We show experimental results where ensemble of smaller ($<$21B) models matches closely 120B MoE (GPT-OSS-120B) on AIME 2025 while larger ensemble of open-weight models reaches 90.0\%, close to proprietary models at time of writing.
	\end{abstract}
]

\printAffiliationsAndNotice{} %

\section{Introduction}
\label{submission}

Mixture-of-Experts \cite{Jacobs91, eigen2014learningfactoredrepresentationsdeep} (MoE) is class of machine learning technique where inputs are routed among multiple expert models sub-modules to produce joint output. With help of Switch Transformers \cite{fedus2022switchtransformersscalingtrillion} routing tokens among expert sub-layers inside one jointly trained model enabled commercial scaling of Large Language Models to trillions of parameters. As pre-training hits diminishing returns the industry has shifted towards Inference-Time Compute Scaling. This shift posits that ``System 2'' thinking that is characterized by deliberate planning and verification that can be synthesized by dynamically allocating resources during generation time \cite{snell2024scaling, deepseekr1, openai2024openaio1card}. Approaches like Titans \cite{behrouz2024titans} and End-to-End Test-Time Training (TTT-E2E) \cite{tandon2025ttt} demonstrate that models can ``learn to memorize'' or update their parameters dynamically on input streams. The Nested Learning paradigm \cite{behrouz2025nested} reframes architectures as hierarchies of optimization problems. Together these suggest that the scaling principles that work neural-networks can be re-applied at higher levels of abstraction, yet composing \emph{separately trained} models and communication between them remains open.

Agentic workflows can be read as the industry to scale in exactly this direction - researchers are composing LLMs into higher-order networks with natural language as the interface. Mixture-of-Agents (MoA) \cite{wang2024mixture} and Chain of Agents (CoA) \cite{zhang2024chain} arrange models as feed-forward acyclic graphs. They solve composability, but lack two distinct properties that make neural networks scale. First, they are not recurrent: layers are not reused with feedback loops as in RNN loops, adding more depth requires more layers, and irreversible upstream error propagation. Second, they have to use central aggregator model. Neural networks aggregate with a fixed, transparent function over ensemble of units; delegating to a single black-box model instead is closer to collapsing a layer into a neuron which not only reduces dimensionality, but, in case of distributed architecture, also has a trusted entity. These systems are further prone to ``herding,'' where a majority of mid-tier agents overpower a lone expert \cite{yi2025debateequilibriumbeliefdrivenmultiagent, cemri2025multiagentllmsystemsfail}, and authority bias \cite{sharma2024understanding}, and degraded context \cite{liu2023lostmiddlelanguagemodels}.
From these limitations we find that \textbf{existing LLM and agentic architectures offer no building block that composes independently trained models into a recurrent layer that can itself be composed recursively}.

In our work we propose Mixture-of-Models (MoM), \textbf {a distributed system architecture that scales reasoning by recurrently applying layers of independently hosted LLMs, whose outputs are combined by a non-linear peer consensus algorithm instead of central aggregator}.

Our approach composes MoM as the N-way self-evaluating deliberation (NSED) protocol: the ensemble evaluates its own outputs, each model scoring only its peers. It consists of the Delphi Method \cite{rescher1998predicting} extended with quadratic voting \cite{lalley2018quadratic} and self-votes masked to damp the herding and authority bias. The resulting consensus state is fed back to the system input. This deterministic procedure produces output as consensus without need for an aggregator model, so the system behaves as a semantic recurrent network. Drawing on sequential analysis and jury theorems \cite{wald1945, condorcet1785essai}, we describe the round-wise accuracy trajectory with a two-parameter model that defines balance between signal extraction and eventual context entropy fatigue; it fits all measured trajectories ($R^2 = 0.81$ - $0.98$) and suggests a stopping round. Voting matrices produce actionable intra-round signal that can indicate divergence for escalation or convergence for early stopping system. All these are useful signals for safety, transparency and human-in-the-loop escalation protocols.
Models exchange in natural language: we deliberately avoid logit or activation layer operations, which would pose architecture matching challenges on vocabulary and scales.
We use \textit{agent} and \textit{model} interchangeably; each agent is one model run in a tool-calling loop.
Our contributions are:

\begin{itemize}
	\item We develop a generic composable distributed inference architecture that enables composing models hosted by independent entities.
	\item We propose a two-parameter model of the accuracy trajectory that fits all runs with $R^2$ 0.81–0.98 and suggests stopping round.
	\item Using proposed design we discuss benefits of construct including human-in-the-loop, explainability and safety features.
	\item We evaluate two ensembles of three open-weight models: a mid-tier ensemble of models of at most 21B parameters (Qwen3-8B, GPT-OSS-20B, Gemma-3-12B) and high-end (Qwen3-Next-80B-A3B, GPT-OSS-120B, DeepSeek-V3.1). The mid tier achieves +48\% on AIME 2025 (53.9\% $\to$ 80.0\%) and by +75\% on LiveCodeBench v5 hard (40.8\% $\to$ 71.2\% Pass@1), and the high-end ensemble by +23\% on AIME 2025 (73.3\% $\to$ 90.0\%), while using convergence based stopping signal ends after 3.8, 2.9 and 2.7 rounds accordingly at average (37-45\% fewer rounds than the best fixed round $T_{opt}$). Ensembles outperformed every individual model on both benchmarks while no single individual model is leading on both. Additionally we demonstrate 37.5\% lower sycophancy and safety features using DarkBench suite \cite{kran2025darkbenchbenchmarkingdarkpatterns}, at the cost of more 'sneaking'" which we argue to be benign protocol artifact.
	\item We make our code publicly available at https://anonymous.4open.science/r/nsed-mom-3454/
\end{itemize}

We posit that MoM represents a recursive scaling of MoE topology - a Scale-Invariant recursion of well established principles in Artificial Intelligence research, transforming multi-agent deliberation from a heuristic art into a composable higher-order reasoning architecture that's paving the road towards scalable Artificial General Intelligence hosted by distributed operators.

\section{Related Work}

Our work synthesizes insights across wide stack of cognitive architecture: from the macro topology of agent graphs and the thermodynamics of inference-time compute, down to the mathematical foundations of high-dimensional consensus. We review these areas to contextualize our departure from existing feed-forward and static architectures.

\paragraph{Topological Limitations in Multi-Agent Systems:} Current Multi-Agent Systems (MAS) are mostly characterized by \textbf{Feed-Forward} information flows. Frameworks like Mixture-of-Agents (MoA) \cite{wang2024mixture} and Microsoft's AutoGen \cite{AutoGen23} structure collaboration as layered DAGs.Chain of Agents (CoA) \cite{zhang2024chain} rely on sequential ``bucket brigade'' pass-offs. These methods suffer from error propagation, where early summarization faults cascade irreversibly downstream. They are also limited to DAG topology that lacks recursive refinement depth.
\paragraph{Inference-Time Compute \& Scale-Invariant Topology:} Nested Learning approach \cite{behrouz2025nested} reframes architectures as hierarchies of optimization problems, meaning that mathematical principles of learning hold true whether the updated module is simple weight, or an entire LLM encapsulated within larger macro level system. The Inference-Time Compute Scaling approach posits that additional compute during generation can substitute for model scale \cite{snell2024scaling, brown2024monkeys}. Titans \cite{behrouz2024titans} and TTT-E2E \cite{tandon2025ttt} implement this by dynamically updating weights or memory buffers during the forward pass. DeepSeek-R1 \cite{deepseekr1} and OpenAI o1 utilize Reinforcement Learning on Chain-of-Thought (CoT) to internalize this deliberation. These methods are limited to model or agent internal topology and don't scale in to \textbf{inter-agent} protocol.

\paragraph{Mechanism Design \& Algorithmic Social Choice:} Mixture-of-Experts (MoE) \cite{fedus2022switchtransformersscalingtrillion} introduced a "gating networks" to route tokens, though it lacks semantic routing and governance layer. From social science direction, Delphi Method \cite{rescher1998predicting} shows deterministic protocol for orchestrating a group of actors to conclude a joint decision, however Xie Yi et.al \cite{yi2025debateequilibriumbeliefdrivenmultiagent, cemri2025multiagentllmsystemsfail} shown that LLM ensembles are vulnerable to `herding'' which limits application of such social choice algorithms. Quadratic Voting \cite{lalley2018quadratic} is another method coming from social sciences shows damping properties that may be useful as countermeasure to herding.

\paragraph{Adaptive Computation \& Cognitive Thermodynamics:}Concern about the energy cost of AI has driven interest in dynamic halting mechanisms \cite{schwartz2020green, graves2016adaptive}. Early works in Adaptive Computation Time (ACT) demonstrated that neural networks can learn to pause processing once a confidence threshold is met. However, these approaches were primarily restricted to microscopic, per-token probability scalars. \textit{Graph of Thoughts} \cite{besta2024graph} or \textit{Tree of Thoughts} \cite{yao2024tree}, optimize for accuracy via static graph expansion. MoM iterates a set of independent models and can stop based on ensemble convergence signal.

\paragraph{Contextual Dynamics \& Routing Efficiency:} \textit{Lost in the Middle}, \cite{liu2023lostmiddlelanguagemodels} showed that while the capacity of context windows has grown, a "U-shaped" attention deficit is common, where models fail to retrieve information located in the middle of long sequences. This phenomenon validates our topological choice of a \textbf{Recurrent State} with a decay factor ($\gamma$) over a linear append-only log. By iteratively compressing history into a refreshed consensus state, our protocol ensures that critical decision boundaries remain within the recency attention head capacity, avoiding the retrieval degradation inherent in standard "Chain of Agents" architectures. Static sparse architectures like \textit{MoE++} \cite{jin2024moeplus} demonstrate that "Zero-Computation Experts" can accelerate inference. MoE++ dynamically drops tokens to save FLOPs, we discuss a Dynamic Expertise Broker that selects experts at the \textit{Session Level}, dynamically dropping entire model instances when their domain utility is low.

\paragraph{Self-Correction \& Bootstrapping:} Recent advances like STaR \cite{zelikman2022star} prove that reasoning traces can serve as high-quality supervision signals for fine-tuning. MoM integrates this insight into the architectural topology itself. By treating the "Consensus State" not just as an output, but as a potential training label for consolidation loop, we bridge the gap between ephemeral Inference-Time Compute \cite{snell2024scaling} and permanent model parameterization.

\paragraph{Theoretical Foundations of Consensus:} To derive our Efficiency-Fatigue Model (Eq. \ref{eq:thermo_governing}), we draw upon foundational results in sequential analysis, geometry, and recent empirical findings on LLM capabilities. Ability to verify that a solution exceeds its ability to generate one zero-shot been empirically established by \citet{cobbe2021training}, who demonstrated that training a verifier scales more effectively than fine-tuning a generator for math problems. \citet{lightman2023lets} showed that "Process Supervision" (step-by-step verification) outperforms outcome-based supervision.

\paragraph{Sequential Analysis:} Wald's Sequential Probability Ratio Test (SPRT) \cite{wald1945} proved that for a sequential sampling process, the decision boundary (consensus) can be reached with minimal observations if the log-likelihood ratio is integrated over time. Our efficiency parameter $\Lambda$ is the topological equivalent of the SPRT information rate. Condorcet's Jury Theorem \cite{condorcet1785essai} proves that the mean verifier precision must strictly exceed random chance ($\bar{p}_v > 0.5$) to guarantee that the collective drift is positive. Without this condition, the sequential integration would accelerate convergence toward hallucination rather than truth.

\paragraph{Geometric Coverage:} The efficacy of heterogeneous ensembles is grounded in high-dimensional geometry. Carathéodory’s Theorem dictates that any point in the convex hull of a $d$-dimensional feature space can be represented by at most $d+1$ vertices. In the context of the \textit{Rashomon Set} of valid reasoning paths \cite{breiman2001rashomon}, this implies that a finite number of diverse agents ($N$) is sufficient to span the solution manifold. Cover's Theorem \cite{cover1965} suggests that projecting a complex classification problem into the high-dimensional space makes the "Truth" linearly separable from "Hallucination".
\section{Theoretical Framework}
\label{sec:architecture}

We propose MoM as a macro-scale analog to Recurrent Neural Networks (RNNs), operating as a neural circuit composed of autonomous agents\footnote{In this work we use words "agent" and "model" interchangeably, assuming that modern LLM applications are mostly agentic.} (Figure \ref{fig:neural_topology}). Following Cover’s Theorem \cite{cover1965} ensemble of models acts as a high-dimensional kernel machine.

While static ensembles are governed by Condorcet's Jury Theorem \cite{condorcet1785essai}, sequential operation is governed by Wald’s Sequential Probability Ratio Test (SPRT). We model the accuracy trajectory $A(t)$ as a thermodynamic competition between signal extraction and entropic fatigue (Eq. \ref{eq:thermo_governing}, full derivation at Appendix \ref{app:derivation}):
\begin{figure}[t]
	\centering
	\setlength{\abovecaptionskip}{0.1in}
	\setlength{\belowcaptionskip}{0.1in}

	\resizebox{\columnwidth}{!}{%
		\begin{tikzpicture}[
				>=stealth,
				node distance=1.2cm,
				font=\small\sf,
				func/.style={rectangle, draw=black, thick, fill=white, minimum height=0.7cm, minimum width=0.9cm, align=center, rounded corners},
				sum/.style={circle, draw=black, thick, fill=white, inner sep=0pt, minimum size=6mm, node contents={+}},
				agent_box/.style={draw=black!80, dashed, thick, fill=white, rounded corners, inner sep=4pt},
				amplifier/.style={isosceles triangle, isosceles triangle apex angle=60, draw=black, thick, fill=white, minimum height=0.6cm, shape border rotate=180, inner sep=2pt, align=center},
				activation/.style={
						rectangle, draw=black, thick, fill=white, minimum size=1.0cm,
						path picture={
								\draw[thick] (-0.35,-0.25) -- (0.35,-0.25);
								\draw[thick] (0,-0.35) -- (0,0.35);
								\draw[thick, red] (-0.35,-0.25) .. controls (0,-0.25) and (0,0.25) .. (0.35,0.25);
							}
					},
				signal/.style={->, thick},
				peer_vote/.style={->, thick, color=blue!70!black},
				self_context/.style={->, dashed, thick, color=gray!80}, %
				feedback/.style={->, thick, dashed, color=red!70!black, rounded corners=5pt},
				control/.style={->, thick, color=blue!70!black, dashed} %
			]

			\node (input) [label=left:$x_t$] {};
			\node (input_sum) [sum, right=0.5cm of input] {};
			\draw[signal] (input) -- (input_sum);

			\node (embed) [func, right=1cm of input_sum, fill=white, minimum height=2.8cm, align=center] {$\mathcal{P}_{sys}$ \\ $S_t$};
			\draw[signal] (input_sum) -- (embed);

			\node (prop1) [func, right=1.2cm of embed, yshift=1.0cm] {$P_1$};
			\node (vote1) [func, right=0.8cm of prop1] {$V_1$};
			\begin{pgfonlayer}{background}
				\node [agent_box, fit=(prop1) (vote1), label=above:\textbf{Model 1}] {};
			\end{pgfonlayer}

			\node (prop2) [func, right=1.2cm of embed] {$P_2$};
			\node (vote2) [func, right=0.8cm of prop2] {$V_2$};
			\begin{pgfonlayer}{background}
				\node [agent_box, fit=(prop2) (vote2)] {};
			\end{pgfonlayer}

			\node (prop3) [func, right=1.2cm of embed, yshift=-1.0cm] {$P_N$};
			\node (vote3) [func, right=0.8cm of prop3] {$V_N$};
			\begin{pgfonlayer}{background}
				\node [agent_box, fit=(prop3) (vote3), label=below:\textbf{Model N}] {};
			\end{pgfonlayer}

			\draw[signal] (embed.east) -- (prop1.west);
			\draw[signal] (embed.east) -- (prop2.west);
			\draw[signal] (embed.east) -- (prop3.west);

			\draw[self_context] (prop1) -- (vote1);
			\draw[self_context] (prop2) -- (vote2);
			\draw[self_context] (prop3) -- (vote3);

			\draw[peer_vote] (prop1.east) -- (vote2.west);
			\draw[peer_vote] (prop1.east) -- (vote3.west);
			\draw[peer_vote] (prop2.east) -- (vote1.west);
			\draw[peer_vote] (prop2.east) -- (vote3.west);
			\draw[peer_vote] (prop3.east) -- (vote1.west);
			\draw[peer_vote] (prop3.east) -- (vote2.west);

			\node (mux) [activation, right=1.2cm of vote2, label=below:{$\sigma_{QV} \odot \mathbf{D}$}] {};

			\draw[signal] (vote1.east) -- (mux.west);
			\draw[signal] (vote2.east) -- (mux.west);
			\draw[signal] (vote3.east) -- (mux.west);

			\node (output) [right=1.2cm of mux, label=right:$y(t)$] {};
			\draw[signal] (mux.east) -- (output);

			\coordinate (tap) at ($(mux.east)!0.5!(output.west)$);
			\node (amp) [amplifier, above=1.6cm of vote2] {$\gamma$};
			\draw[feedback] (tap) -- ++(0, 1.6) |- (amp.east);
			\draw[feedback] (amp.west) -| (input_sum.north);

		\end{tikzpicture}%
	}
	\caption{The NSED layer. \textbf{Solid Blue Lines} indicate actionable voting signals, while \textbf{Dashed Gray Lines} indicate read-only context. The instantaneous output $y(t)$ is appended to the Deliberation History $\mathcal{H}$, which is subsequently processed by the Weighted Consensus function (Eq. \ref{eq:weighted_consensus}) to derive the final solution $y^*$.}
	\label{fig:neural_topology}
\end{figure}

\begin{equation}
	\label{eq:thermo_governing}
	\text{A}(t) = \underbrace{1 - (1 - p_g)e^{-\Lambda (p_v - p_g) (t-1)}}_{\text{Signal Extraction}} - \underbrace{\beta (t-1)^2}_{\text{Context Fatigue}}
\end{equation}
Where $t$ is the round, ($t \in \{1, 2, \dots, T_{max}\}$) where $T_{max}$ is the maximum number of rounds allowed; $p_g$ is the ensemble's first round ($t=1$) generation accuracy, $p_v$ is the mean verification accuracy of evaluators. Convergence requires $p_v > p_g$ ; $\Lambda$ is process efficiency and $\beta$ is the fatigue coefficient, representing the accumulation of context noise (entropy) over time. Equation \ref{eq:thermo_governing} is thus a two-parameter model whose maximum performance $T_{opt}$ estimates round where accuracy of the ensemble peaks; $T_{opt}$ itself can be obtained from calibration datasets.

\begin{figure}[t]
	\centering
	\setlength{\abovecaptionskip}{0.1in}
	\setlength{\belowcaptionskip}{0.1in}

	\resizebox{\columnwidth}{!}{%
		\begin{tikzpicture}[
				node distance=0.8cm and 1cm,
				auto,
				font=\sf\small,
				block/.style={rectangle, draw, thick, rounded corners, minimum height=2em, text centered, fill=white},
				cloud/.style={ellipse, draw, thick, fill=white, node distance=1.5cm, minimum height=2em},
				env/.style={cylinder, shape border rotate=90, draw, thick, aspect=0.25, fill=white, minimum height=1em, minimum width=1.5em},
				line/.style={draw, -latex', thick},
				data/.style={draw, -latex', dashed, thick, color=blue!70!black}
			]

			\node[block] (orchestrator) {\textbf{Orchestrator}};
			\node[cloud, above=0.8cm of orchestrator] (memory) {NSED State $S_t$};
			\node[block, left=2cm of memory] (broker) {\textbf{Broker}};
			\node[block, fill=white, below=0.8cm of broker] (user) {Client / Task};

			\node[block, right=0.5cm of orchestrator] (agent1) {Agent 1};
			\node[block, below=0.25cm of agent1] (agent2) {Agent ...};
			\node[block, below=0.25cm of agent2] (agent3) {Agent N};

			\node[env, align=center, right=1cm of memory] (oracle) {\textbf{Environment}\\(Oracle/Files)};

			\draw[line] (user) -- node[above, sloped, font=\scriptsize] {Task \& SLA} (orchestrator);
			\draw[line] (orchestrator) -- node[left, font=\scriptsize] {Get Agents} (broker);
			\draw[line] (user) -- node[left, font=\scriptsize] {SLA Inquiry} (broker);
			\draw[line] (orchestrator) -- (memory);

			\draw[line] (orchestrator.east) |- (agent1.west);
			\draw[line] (orchestrator.south) |- (agent2.west);
			\draw[line] (orchestrator.south) |- (agent3.west);

			\draw[data] (agent1.east) -| (oracle.south);
			\draw[data] (agent2.east) -| (oracle.south);
			\draw[data] (agent3.east) -| node[below, pos=0.5, font=\scriptsize, color=blue!70!black, sloped] {Data Shards} (oracle.south);

			\draw[data] (oracle.south) -- ++(0,-4cm) -| (user.south)
			node[pos=0.25, below, font=\tiny, color=blue!70!black] {Direct Client Inquiries};

		\end{tikzpicture}%
	}
	\caption{Control flow diagram representing agentic interpretation. Agents maintain two distinct interfaces: a Protocol Interface to the Orchestrator for peer-review, and a Side-Channel to the Environment for independent context retrieval is equivalent of models being trained upon different datasets. Each agent represents a model within ReAct agent loops harness.}
	\label{fig:control_flow_agent_representation}
\end{figure}

System control plane is demonstrated at Figure \ref{fig:control_flow_agent_representation} and shows how model level thinking (Figure \ref{fig:neural_topology}) can be interpret as agentic harness definitions, broker and general optimisation framework is detailed at Appendix \ref{app:opt_fw}

\subsection{N-Way Self-Evaluating Deliberation (NSED)}\label{sec:neural_circuit}
Just as a standard recurrent neural network projects inputs into a hidden state, we project the raw task $x_t$ and the prior System State $S_{t-1}$ into the semantic space of the agents via the system prompt ($\mathcal{P}_{sys}$).

\begin{equation}
	h_t = \text{Embedding}\left( [x_t ; S_{t-1}] \mid \mathcal{P}_{sys} \right)
\end{equation}

Where $t$ is current deliberation Round Index ($t=1 \dots T_{max}$); $S_{t}$ is system state containing the aggregated proposals, evaluations, and reasoning traces from previous rounds; $[ \cdot ; \cdot ]$ is semantic concatenation of the dynamic input buffer and the recurrent state.

Forward pass takes natural language input $h_t$ that's duplicated $N$ times and fed into the model pool along the $\mathcal{P}_{sys}$. Each model $M_i$ processes the input in parallel, generating a candidate $P_i$. Each agent gets input for round $t$ with a summary of round $t-1$: first characters of each proposal (ranked by score) and critiques it received, full peer scores table. Older rounds reach agents via scratchpad filled on demand through history tools (read\_proposal, read\_critiques)

These candidates then are passed to hidden evaluation layer, they are passed with candidate names that do not link back to model name and self-voting option removed (Appendix \ref{app:topology_blinding}) and cross-evaluated by voting layer.

The Voting Layer ($V_i$ in Figure \ref{fig:neural_topology}) aggregates the proposal content ($P_i$) and assigns peer-weights ($v$) to enable selecting the leading state candidate.  Let $\vec{v}_j$ be the votes vector evaluator $j$ issues in a round, masked by self-vote mask $\mathbf{D}$ ($0$ when $i=j$). The design mitigates possible attempts by agents to game system, self-vote or hallucinates by giving out-of-budget scores ($\lVert \vec{v}_{j} \rVert_1 > B$) due to clamp and $\mathbf{D}$ self-vote mask:

\begin{equation}
       \label{eq:self_mask}
       \vec{v}_{j} = (V_t \odot \mathbf{D})_j
 \end{equation}

 The votes are then normalised to the budget:
\begin{equation}
      \label{eq:vote_clamp}
      \hat{\vec{v}}_{j} = \frac{\vec{v}_{j}}{\max\left(B,\ \lVert \vec{v}_{j} \rVert_1\right)}
\end{equation}

  Where $B$ is the voting weight budget of each evaluator ($B = 100$ in our runs).

  \begin{equation}
        \label{eq:qv_activation}
        \sigma_{QV}(V_t) = \sqrt{\hat{V}_t}
  \end{equation}

Where $\hat{V}_t = (\hat{\vec{v}}_1, \dots, \hat{\vec{v}}_N)^\top$ stacks the normalised votes as rows, and the square root is taken entry-wise.

\paragraph{The Feedback Loop:}
\label{par:feedback_loop}

The global context provided to agents is the concatenation $[x_t ; S_{t-1}]$. where $x_t$ is the cumulative user input append-only buffer ($x_t = x_{t-1} \cup u_t$). If a user injects new data $u_t$ (e.g., "add this constraint") between rounds, it updates the boundary conditions for all subsequent iterations. State ($S_t$) is the evolving solution. This is mutable and subject to the voting function $\sigma_{QV}$. The state update rule focuses strictly on integrating reasoning:
\begin{equation}
	S_t = (\gamma_t \odot S_{t-1}) + \sigma_{QV}(V_t) \odot P_t
\end{equation}
Where: $\gamma_t$ acts as an attention regulator. A high $\gamma$ maintains the prior consensus (refinement), while a low $\gamma$ offloads prior states (pivoting).

Upon termination, the final output $y^*$ is derived via a \textit{Weighted Consensus} function over the entire deliberation history $\mathcal{H}$. Rather than relying solely on the final state, which may suffer from local regression, we generalize system to support the time-weighted influence score:

\begin{equation}\label{eq:weighted_consensus}
	y^* = \operatorname*{argmax}_{p \in \mathcal{H}} \left( s(p) \cdot \omega(t_p) \right)
\end{equation}

Where $p$ is all proposals in the history $\mathcal{H}$, $t_p$ is the round of $p$, $s(p) = \frac{1}{|E_p|}\sum_{j \in E_p} \sigma_{QV}(V_t)_{j,p})$ is the mean QV score (Eq.~\ref{eq:qv_activation}) that $p$ received from its evaluators $E_p$, and $\omega(t)$ is a temporal weighting kernel.

This abstraction supports diverse policies ranging from last round best ($\omega(t) = \delta_{t,T_{max}}$) to treating the final consensus as a trajectory optimization problem. Specific instantiations of $\omega(t)$ are detailed in Appendix \ref{app:setup:cons_strat}.

\subsection{Convergence based stopping}\label{sec:stopping}
Equation \ref{eq:thermo_governing} predicts the best round, however it would be very wasteful to iterate until $T_{opt}$ in trivial task cases. We derive a stopping signal from voting layer itself. Under quadratic voting budget (Eq. \ref{eq:vote_clamp}), a complete over a board agreement happens when every evaluator scoring $n$ equally, delivering $\sqrt{n}$ of effective power. Thus we define convergence score as:
\begin{equation}
       \label{eq:convergence}
       C(t) = \frac{\sum_j \lVert \sigma_{QV}(V_t)_j \rVert_1}{\sum_j \sqrt{n_j(t)}} \in [0, 1],
 \end{equation}
 where $j$ denotes row of Eq.~\ref{eq:qv_activation}.
Accuracy-weighted convergence is then
\begin{equation}
	\label{eq:weighted_convergence}
	C_{weighted}(t) = C(t) \cdot A_{norm}(t),
\end{equation}
where $A_{norm}(t)$ is Equation~\ref{eq:thermo_governing} divided by its peak value. Deliberation stops at the first round with $C_{weighted}(t) \ge \theta$, where $\theta \in [0, 1]$ is a threshold hyperparameter.

\section{Experimental Methodology}
\label{sec:methodology}
\begin{algorithm}[t]
\caption{NSED with convergence-based stopping and broker escalation. An answer $y(t)$ is available after every round. Runs used fixed $T_{max}$; stopping and escalation were applied offline to the recorded rounds (Sections~\ref{sec:stopping} and~\ref{sec:broker}). Failed agents are retried, then skipped (Appendix~\ref{app:failures}).}
\label{alg:nsed_protocol}
\begin{algorithmic}[1]
\State \textbf{Input:} Task $x$, Ensembles $\mathcal{A}_{low}, \mathcal{A}_{high}$, Requirement $\tau$, Threshold $\theta$, Budget $B$, Rounds $T_{max}$
\State \textbf{Output:} Solution $y^*$, Intermediate answers $y(t)$
\Function{Round}{$x, \mathcal{A}, S_{t-1}$}
	\State $P_{k,t} \leftarrow M_k.\text{propose}(x, S_{t-1})$ \textbf{for all} $M_k \in \mathcal{A}$
	\State $V_t \leftarrow$ votes of every $M_j$ on $\{P_{i,t}\}_{i \neq j}$
	\State $\vec{v}_j \leftarrow (V_t \odot \mathbf{D})_j$ \Comment{Eq.~\ref{eq:self_mask}}
	\State $\hat{\vec{v}}_j \leftarrow \vec{v}_j / \max(B, \lVert \vec{v}_j \rVert_1)$ \Comment{Eq.~\ref{eq:vote_clamp}}
	\State $\sigma_{QV}(V_t) \leftarrow \sqrt{\hat{V}_t}$ \Comment{Eq.~\ref{eq:qv_activation}}
	\State $s_{i,t} \leftarrow$ mean of column $i$ over its evaluators, or $-1$
	\State $y(t) \leftarrow P_{i^*,t}$, $i^* = \arg\max_i s_{i,t}$
	\State \Return $y(t)$, $C(t)$, $S_t \leftarrow \text{Summary}(t)$ \Comment{Eq.~\ref{eq:convergence}}
\EndFunction
\Statex
\State $\mathcal{A} \leftarrow \mathcal{A}_{low}$, $S_0 \leftarrow \emptyset$, $t \leftarrow 1$
\While{$t \le T_{max}$}
	\State $y(t), C(t), S_t \leftarrow \Call{Round}{x, \mathcal{A}, S_{t-1}}$
	\State $y^* \leftarrow y(t)$ \Comment{current answer, available every round}
	\If{$C(t) \cdot A_{norm}(t) \ge \theta$} \Comment{Eq.~\ref{eq:weighted_convergence}}
		\State \textbf{break}
	\ElsIf{$\mathcal{A} = \mathcal{A}_{low}$ \textbf{and} $t = 2$ \textbf{and} $C(2) < \tau$}
		\State $\mathcal{A} \leftarrow \mathcal{A}_{high}$, $S_0 \leftarrow \emptyset$, $t \leftarrow 1$ \Comment{escalate}
	\Else
		\State $t \leftarrow t + 1$
	\EndIf
\EndWhile
\State \Return $y^*$
\end{algorithmic}
\end{algorithm}
The execution logic of the protocol is detailed in Algorithm \ref{alg:nsed_protocol}. We evaluate it on AIME 2025 and on the hard subset of LiveCodeBench v5. Two ensembles are used. The mid-tier ensemble (Qwen3-8B, GPT-OSS-20B, Gemma-3-12B) and the high-end ensemble (Qwen3-Next-80B-A3B, GPT-OSS-120B, DeepSeek-V3.1). Each deliberation was set to run for a fixed number of rounds (7 mid-tier, 8 high-end) following Algorithm \ref{alg:nsed_protocol}. The state $S_{t-1}$ of the update rule is passed to every agent as summary of round $t-1$: every candidate's answer first 300 characters, with author and peer scores ranked by score agents own $t-1$ generated content and peer critiques it received. Older rounds enter context only through agent's scratchpad filled by calling tools. This design aimed to represent a fixed $\gamma$ which modulates attention towards previous round data. We report accuracy at convergence-based stopping (Eq. \ref{eq:weighted_convergence}, $\theta = 0.9$ for AIME, $0.7$ for LiveCodeBench). The stopping round is chosen during runtime from the convergence signal, not after the fact (Section~\ref{sec:stopping}). We measure $p_g$ and $p_v$ from the recorded runs and fit $\Lambda$ and $\beta$ per configuration (detailed in Appendix Table~\ref{tab:fit}). Beyond reasoning performance, we evaluated the protocol's ability to mitigate manipulative behaviors using the DarkBench suite \cite{kran2025darkbenchbenchmarkingdarkpatterns}.

All baselines we report were re-measured with our harness, using base agent prompt, budget and temperature. We report each member on its own and the average member the mean of members accuracies. On AIME, each mid-tier member answers by majority over three samples elsewhere each member answers once.
Detailed sampling and serving settings are listed in Appendix~\ref{app:setup}.

\section{Results}
Results obtained for AIME2025 are shown in Figure \ref{fig:main_aime_figure}. MoM ensembles fixed for stopping at $T_{opt}$ achieved 82.5\% precision for the mid-tier ensemble and 91.1\% for the high-end ensemble, resulting in \textbf{+53\% and +24\% relative gain} over average member accordingly. On LiveCodeBench (v5, hard subset) shown in Figure \ref{fig:lcb_trajectory}, same mid-tier ensemble reached Pass@1 of \textbf{71.8\%} at $T_{opt}$, resulting \textbf{+76\% relative gain in coding performance}. Results are summarised in Table \ref{tab:global_comparison}. The observed trajectories align with our empirical model (Eq. \ref{eq:thermo_governing}). The fit ($R^2 = 0.98$ and $0.81$) is consistent with the late-round decline being predictable system property. When combined with convergence based stopping (Equation \ref{eq:weighted_convergence}, $\theta = 0.9$ for AIME, $0.7$ for LiveCodeBench) MoM reached 80.0\%, 90.0\% and 71.2\% (\textbf{+48\%}, \textbf{+23\%} and \textbf{+75\%} over the average member) while stopping after 3.8, 2.7 and 2.9 rounds on average instead of $T_{opt}$ = 6, 5 and 5 (Table~\ref{tab:stopping_matrix}); the full $\theta$ sweep is given in Figure~\ref{fig:stopping_tradeoff}.
In our runs Qwen3-8B missed 11.2\% of its AIME proposal turns and 37.2\% of its LiveCodeBench evaluations (Table~\ref{tab:failures}), and the ensemble still reached the reported accuracy, demonstrating reliable failure-mode handling (Appendix Table~\ref{tab:failures}).

\begin{figure}[t]
   \centering
   \begin{minipage}{\linewidth}
       \centering
       \begin{tikzpicture}
           \begin{axis}[
                   width=0.97\linewidth, height=0.7\linewidth, %
                   xlabel={\textbf{Round ($t$)}},
                   ylabel={\textbf{Pass@1 Score}},
                   xmin=1, xmax=7,
                   ymin=0.4, ymax=1.0,
                   legend pos=south east,
                   legend style={font=\tiny},
                   grid=both,
                   thick
               ]
               \addplot[color=green!60!black, mark=*, line width=1.2pt]
               coordinates {(1,0.692) (2,0.742) (3,0.775) (4,0.808) (5,0.817) (6,0.825) (7,0.792)};
               \addlegendentry{MoM (measured)}
               \addplot[domain=1:7, samples=50, color=blue!80!black, dashed, line width=1pt]
               {1 - (1-0.692)*exp(-4.95*0.041*(x-1)) - 0.0030*(x-1)^2};
               \addlegendentry{Eq.~\ref{eq:thermo_governing} fit ($R^2{=}0.98$)}
               \addplot[color=orange!80!black, dashdotted, line width=1pt] coordinates {(1,0.742) (7,0.742)};
               \addlegendentry{Best member SC@3 (74.2\%)}
               \addplot[color=red!80!black, dotted, line width=1pt] coordinates {(1,0.539) (7,0.539)};
               \addlegendentry{Average member (53.9\%)}

           \end{axis}
       \end{tikzpicture}

       \vspace{0.05in}
       {\small (a) Mid-Tier setup ($\Lambda=4.95, \beta=0.0030$).}
   \end{minipage}

   \vspace{0.15in} %

   \begin{minipage}{\linewidth}
       \centering
       \begin{tikzpicture}
           \begin{axis}[
                   width=0.97\linewidth, height=0.7\linewidth,
                   xlabel={\textbf{Round ($t$)}},
                   ylabel={\textbf{Pass@1 Score}},
                   xmin=1, xmax=7,
                   ymin=0.6, ymax=1.0,
                   legend pos=south east,
                   legend style={font=\tiny},
                   grid=both,
                   thick
               ]
               \addplot[color=green!60!black, mark=*, line width=1.2pt]
               coordinates {(1,0.822) (2,0.889) (3,0.889) (4,0.889) (5,0.911) (6,0.911) (7,0.900) (8,0.900)};
               \addlegendentry{MoM (measured)}
               \addplot[domain=1:8, samples=50, color=blue!80!black, dashed, line width=1pt]
               {1 - (1-0.822)*exp(-8.65*0.030*(x-1)) - 0.0016*(x-1)^2};
               \addlegendentry{Eq.~\ref{eq:thermo_governing} fit ($R^2{=}0.81$)}
               \addplot[color=orange!80!black, dashdotted, line width=1pt] coordinates {(1,0.811) (8,0.811)};
               \addlegendentry{Best member (81.1\%)}
               \addplot[color=red!80!black, dotted, line width=1pt] coordinates {(1,0.733) (8,0.733)};
               \addlegendentry{Average member (73.3\%)}
       \end{axis}
       \end{tikzpicture}

       \vspace{0.05in}
       {\small (b) High-End setup ($\Lambda=8.65, \beta=0.0016$).}
   \end{minipage}

   \setlength{\abovecaptionskip}{0.1in}
   \setlength{\belowcaptionskip}{0.1in}

   \caption{Ensemble performance overlaid with thermodynamic equation fit curves.}
   \label{fig:main_aime_figure}
	\end{figure}
	\begin{figure}[t]
    \centering
    \begin{tikzpicture}
        \begin{axis}[
                width=0.97\columnwidth, height=0.8\columnwidth,
                xlabel={\textbf{Deliberation Round ($t$)}}, ylabel={\textbf{Pass@1}},
                xmin=1, xmax=7, ymin=0.10, ymax=0.80,
                ytick={0.10,0.20,0.30,0.40,0.50,0.60,0.70,0.80},
                yticklabels={10\%,20\%,30\%,40\%,50\%,60\%,70\%,80\%},
                legend pos=south east, legend style={font=\tiny},
                grid=both, grid style={line width=0.5pt, draw=black!20},
                major grid style={line width=0.5pt, draw=black!40}, thick]
            \addplot[color=green!60!black, mark=*, line width=1.2pt]
            coordinates {(1,0.619) (2,0.683) (3,0.715) (4,0.701) (5,0.718) (6,0.703) (7,0.709)};
            \addlegendentry{MoM (measured)}
            \addplot[domain=1:7, samples=50, color=blue!80!black, dashed, line width=1pt]
            {1 - (1-0.619)*exp(-10.25*0.0140*(x-1)) - 0.0039*(x-1)^2};
            \addlegendentry{Eq.~\ref{eq:thermo_governing} fit ($R^2{=}0.84$)}
            \addplot[color=orange!80!black, dashdotted, line width=1pt] coordinates {(1,0.529) (7,0.529)};
            \addlegendentry{Best member, Gemma-3-12B (52.9\%)}
            \addplot[color=red!80!black, dashed, line width=1pt] coordinates {(1,0.408) (7,0.408)};
            \addlegendentry{Average member (40.8\%)}
        \end{axis}
    \end{tikzpicture}
    \caption{LiveCodeBench v5 hard ($n{=}344$), mid-tier ensemble. Dashed: Eq.~\ref{eq:thermo_governing} with measured $p_g{=}0.619$, $p_v{=}0.633$ and fitted $\Lambda{=}10.2$, $\beta{=}0.0039$ ($R^2{=}0.84$, $t_{opt}{\approx}4.9$).}
    \label{fig:lcb_trajectory}
	\end{figure}

\paragraph{Safety \& Alignment:} Mid-tier ensemble demonstrated \textbf{+37.5\% sycophancy reduction over the best ensemble model} and overall \textbf{Group Mean improvement up to 10.88\%} when base model mean value is compared with 3 rounds of deliberation. DarkBench judge has constantly scored MoM with high "Sneaking" rating, which we attribute to benign NSED deliberation artifacts (i.e. \textit{"Critics integration:..."}) leaking in to final answers. Detailed DarkBench safety alignment suite is given in Appendix \ref{app:darkbench}.

\subsection{Broker Telemetry \& Influence Dynamics}\label{sec:broker_empiric}

The temporal analysis (Figure \ref{fig:leaders_across_rounds}) shows that there seems to be connection between voting volatility and performance breakthroughs. While initial round distribution depends on ability to generate correct solution, in subsequent rounds models ability to learn also matters, eventually followed by ability to consolidate at peak accuracy rounds and eventually to sustain context fatigue at the last round. Figure \ref{fig:social_graph} also reveals that evaluation and voting patterns generally hold truth, with even weak generators showing evaluation capabilities.
   	\begin{figure}[h]
       \centering
       \pgfplotsset{colormap={whitered}{color=(white) color=(red)}}
       \begin{tikzpicture}
           \begin{axis}[
                   width=0.65\columnwidth, height=3.5cm, scale only axis,
                   enlargelimits=false, axis on top, view={0}{90},
                   tick label style={font=\tiny}, major tick length=0pt,
                   xlabel={\textbf{Round}},
                   xtick={0,1,2,3,4,5,6},
                   xticklabels={R1,R2,R3,R4,R5,R6,R7},
                   ytick={0,1,2,3,4,5},
                   yticklabels={
                           Qwen3-Next-80B, GPT-OSS-120B, DeepSeek-V3.1, %
                           Qwen3-8B, GPT-OSS-20B, Gemma-3-12B           %
                       },
                   colormap name=whitered, point meta min=0, point meta max=1,
                   colorbar, colorbar style={font=\tiny, width=0.2cm, major tick length=0pt},
                   nodes near coords,
                   nodes near coords style={anchor=center, font=\tiny, inner sep=0pt, /pgf/number format/fixed, /pgf/number format/fixed zerofill, /pgf/number format/precision=2},
                   thick
               ]
               \addplot[matrix plot*, mesh/cols=7, point meta=explicit]
               coordinates {
                       (0,0) [0.28] (1,0) [0.13] (2,0) [0.20] (3,0) [0.21] (4,0) [0.24] (5,0) [0.17] (6,0) [0.17]
                       (0,1) [0.53] (1,1) [0.73] (2,1) [0.72] (3,1) [0.72] (4,1) [0.66] (5,1) [0.73] (6,1) [0.72]
                       (0,2) [0.19] (1,2) [0.13] (2,2) [0.08] (3,2) [0.07] (4,2) [0.10] (5,2) [0.10] (6,2) [0.11]
                       (0,3) [0.14] (1,3) [0.33] (2,3) [0.36] (3,3) [0.37] (4,3) [0.24] (5,3) [0.25] (6,3) [0.28]
                       (0,4) [0.62] (1,4) [0.52] (2,4) [0.49] (3,4) [0.48] (4,4) [0.52] (5,4) [0.57] (6,4) [0.46]
                       (0,5) [0.24] (1,5) [0.16] (2,5) [0.15] (3,5) [0.15] (4,5) [0.24] (5,5) [0.17] (6,5) [0.26]
                   };
               \draw[black, dashed, line width=1.5pt] (axis cs:-0.5,2.5) -- (axis cs:6.5,2.5);
           \end{axis}
       \end{tikzpicture}

       \setlength{\abovecaptionskip}{0.1in}
       \setlength{\belowcaptionskip}{0.1in}

       \caption{Share of rounds won by each agent on AIME 2025 (top three rows: mid-tier ensemble; bottom three: high-end). Notice winrate distribution changes between round 1 and 2 where initial ability to learn impacts and at rounds 5-6 where $T_{opt}$ consolidation is.}
       \label{fig:leaders_across_rounds}
	\end{figure}

	\begin{table}[t]
    \centering
    \caption{Accuracy on AIME 2025 and LiveCodeBench v5 (hard). MoM: convergence-based stopping (Eq.~\ref{eq:weighted_convergence}, $\theta=0.9$ for AIME, $0.7$ for LiveCodeBench). Baselines re-measured with our harness (\S\ref{sec:methodology}).}
    \label{tab:global_comparison}
    \resizebox{\columnwidth}{!}{%
    \begin{tabular}{l c c}
        \toprule
        \textbf{System} & \textbf{AIME 2025} & \textbf{LCB v5 hard} \\
        \midrule
        \multicolumn{3}{l}{\textit{Mid-tier members (Qwen3-8B, GPT-OSS-20B, Gemma-3-12B)}} \\
        GPT-OSS-20B              & 74.2\%        & 36.6\% \\
        Qwen3-8B                 & 64.2\%        & 32.8\%$^\ddagger$ \\
        Gemma-3-12B              & 23.3\%        & 52.9\% \\
        Average member           & 53.9\%        & 40.8\% \\
        \textbf{MoM (mid-tier)} & \textbf{80.0\%}  & \textbf{71.2\%} \\
        \midrule
        \multicolumn{3}{l}{\textit{High-end members (Qwen3-Next-80B, GPT-OSS-120B, DeepSeek-V3.1)}} \\
        GPT-OSS-120B             & 81.1\%        & -- \\
        Qwen3-Next-80B           & 78.9\%        & -- \\
        DeepSeek-V3.1            & 60.0\%        & -- \\
        Average member           & 73.3\%        & -- \\
        \textbf{MoM (high-end)} & \textbf{90.0\%}  & --$^\dagger$ \\
        \bottomrule
    \end{tabular}}
    \par\vspace{2pt}{\footnotesize Mid-tier AIME members: majority over 3 samples; all others: one sample.
    $^\ddagger$Qwen3-8B returned no answer for 204 of 344 problems when queried alone (empty or error) and is scored 0 on them; inside MoM its round-1 proposals pass 53.2\%.
    $^\dagger$High-end ensemble evaluated on AIME only.}
	\end{table}

	\begin{figure}[t]
    \centering
    \pgfplotsset{colormap={whitered}{color=(white) color=(red)}}

    \begin{minipage}{\linewidth}
        \centering
        \begin{tikzpicture}
            \begin{axis}[
                    width=0.55\linewidth, height=0.25\linewidth,
                    enlargelimits=false, axis on top, view={0}{90},
                    tick label style={font=\tiny},
                    xlabel={\textbf{Generator}}, ylabel={\textbf{Evaluator}},
                    xlabel style={yshift=10pt},
                    ylabel style={yshift=15pt}, %
                    xtick={0,1,2}, xticklabels={Gemma-12B, GPT-OSS-20B, Qwen3-8B},
                    ytick={0,1,2}, yticklabels={Qwen3-8B, GPT-OSS-20B, Gemma-12B},
                    major tick length=0pt,
                    x tick label style={rotate=0, anchor=north, inner sep=0pt, yshift=-2pt},
                    y tick label style={rotate=0, inner sep=0pt, xshift=-2pt},
                    colormap name=whitered, point meta min=0, point meta max=1,
                    nodes near coords, scale only axis,
                    nodes near coords style={anchor=center, font=\footnotesize, inner sep=0pt, /pgf/number format/fixed, /pgf/number format/precision=2},
                    thick
                ]
                \addplot[matrix plot*, mesh/cols=3, point meta=explicit]
                coordinates {
                        (0,0) [0.51] (1,0) [0.74] (2,0) [0.00]
                        (0,1) [0.48] (1,1) [0.0] (2,1) [0.71]
                        (0,2) [0.00] (1,2) [0.67] (2,2) [0.61]
                    };
            \end{axis}
        \end{tikzpicture}

        \vspace{0.05in}
        {\small (a) Consumer-Grade Ensemble Influence}
    \end{minipage}

    \vspace{0.15in} %

    \begin{minipage}{\linewidth}
        \centering
        \begin{tikzpicture}
            \begin{axis}[
                    width=0.55\linewidth, height=0.25\linewidth,
                    enlargelimits=false, axis on top, view={0}{90},
                    tick label style={font=\tiny},
                    xlabel={\textbf{Generator}}, ylabel={\textbf{Evaluator}},
                    xlabel style={yshift=10pt},
                    ylabel style={yshift=15pt}, %
                    xtick={0,1,2}, xticklabels={DeepSeek, GPTOSS-120B, Qwen3-80B},
                    ytick={0,1,2}, yticklabels={Qwen3-80B, GPTOSS-120B, DeepSeek},
                    major tick length=0pt,
                    x tick label style={rotate=0, anchor=north, inner sep=0pt, yshift=-2pt},
                    y tick label style={rotate=0, inner sep=0pt, xshift=-2pt},
                    colormap name=whitered, point meta min=0, point meta max=1,
                    nodes near coords, scale only axis,
                    nodes near coords style={anchor=center, font=\footnotesize, inner sep=0pt, /pgf/number format/fixed, /pgf/number format/precision=2},
                    thick
                ]
                \addplot[matrix plot*, mesh/cols=3, point meta=explicit]
                coordinates {
                        (0,0) [0.65] (1,0) [0.65] (2,0) [0.00]
                        (0,1) [0.67] (1,1) [0.0] (2,1) [0.71]
                        (0,2) [0.00] (1,2) [0.73] (2,2) [0.63]
                    };
            \end{axis}
        \end{tikzpicture}

        \vspace{0.05in}
        {\small (b) High-Performance Ensemble Influence}
    \end{minipage}

    \setlength{\abovecaptionskip}{0.1in}
    \setlength{\belowcaptionskip}{0.1in}

    \caption{Broker Telemetry: Inter-Agent Influence Matrices. The asymmetry (strong columns vs. distributed rows) indicates that while specific models dominate generation, evaluation is a distributed burden, validating the heterogeneous allocation strategy.}
    \label{fig:social_graph}
	\end{figure}

	\paragraph{Broker emulation.}\label{sec:broker} The same convergence signal can drive expert selection. On AIME, the convergence score $C(2)$ of the mid-tier ensemble predicts whether that ensemble's answer  at the stopping round will be correct (AUC 0.80; 0.77 after round~1). We emulate a broker that starts every question on the mid-tier ensemble and, if $C(2) < \tau$, discards that run and answers the question with the high-end ensemble. Both ensembles use the same stopping rule of Section~\ref{sec:stopping}.
	Table~\ref{tab:broker} shows that escalating the third of questions on which the mid-tier ensemble converges weakest recovers 7.2 of the 10pp separating the two ensembles, 4pp more than escalating same questions at random. As with token-level routing in Mixture-of-Experts, the routing signal comes from the system internal data. The emulation discards the escalated mid-tier deliberation context for sake of test harness simiplicity; this is clear future improvement opportunity.

	\begin{table}[ht]
    \centering
    \caption{Broker emulation on AIME 2025: escalate from the mid-tier to the high-end ensemble when $C(2)$ is below $\tau$. Random column compares precision gains at escalation corresponding random probability}
    \label{tab:broker}
    \small
    \begin{tabular}{l c c c}
        \toprule
        \textbf{Policy} & \textbf{Escalated} & \textbf{Accuracy} & \textbf{Random} \\
        \midrule
        Mid-tier only            & 0\%   & 80.0\% & -- \\
        Escalate, $\tau = 0.6$   & 16\%  & 83.3\% & 81.6\% \\
        Escalate, $\tau = 0.7$   & 32\%  & 87.2\% & 83.2\% \\
        Escalate, $\tau = 0.8$   & 44\%  & 88.6\% & 84.4\% \\
        Escalate, $\tau = 0.9$   & 57\%  & 89.7\% & 85.8\% \\
        High-end only            & 100\% & 90.0\% & -- \\
        \bottomrule
    \end{tabular}
	\end{table}

\subsection{Ablation Studies}\subsubsection{Quadratic Voting}
We isolated the impact of Equation \ref{eq:qv_activation}, the quadratic voting budget. AIME'25 results (Figure \ref{fig:combined_ablations}) show that without it the ensemble falls behind with best round down to 78.3\% versus 82.5\%.

\subsubsection{Model Heterogeneity}
To verify that accuracy improvements stem from model heterogeneity rather than inference-time scaling we compared exactly same protocol executed on a best performing model of both mid-tier and high-end ensembles, results confirm our intuition and are shown in Table \ref{tab:homogeneous_ablation}. (Full trajectory data is provided in Appendix \ref{app:ablation})

\begin{figure}[t]
    \centering
    \begin{tikzpicture}
        \begin{axis}[
                width=0.97\linewidth, height=0.65\linewidth,
                xlabel={\textbf{Deliberation Round}},
                ylabel={\textbf{Accuracy (AIME 2025)}},
                xmin=1, xmax=7,
                ymin=0.6, ymax=0.9,
                xtick={1,2,3,4,5,6,7},
                ytick={0.6, 0.7, 0.8, 0.9},
                yticklabels={60\%,70\%,80\%,90\%},
                legend pos=south east,
                legend style={font=\tiny, fill=white, fill opacity=0.9},
                grid=both,
                grid style={line width=0.5pt, draw=black!15},
                major grid style={line width=0.5pt, draw=black!30},
                thick
            ]
            \addplot[color=green!60!black, mark=*, line width=1.2pt]
            coordinates {(1,0.692) (2,0.742) (3,0.775) (4,0.808) (5,0.817) (6,0.825) (7,0.792)};
            \addlegendentry{Full protocol}
            \addplot[color=red!80!black, mark=square*, line width=1.2pt]
            coordinates {(1,0.683) (2,0.742) (3,0.742) (4,0.708) (5,0.758) (6,0.758) (7,0.783)};
            \addlegendentry{No vote budget}
        \end{axis}
    \end{tikzpicture}

    \setlength{\abovecaptionskip}{0.1in}
    \setlength{\belowcaptionskip}{0.1in}

    \caption{Quadratic-voting ablation, AIME 2025, mid-tier ensemble: accuracy per round with and without the vote budget.}
    \label{fig:combined_ablations}
\end{figure}

\begin{table}[hb]
    \centering
    \caption{AIME 2025 accuracy, homogeneous vs.\ heterogeneous ensembles (best round). Homogeneous high-perf ran 5 rounds, heterogeneous 8.}
    \label{tab:homogeneous_ablation}
    \resizebox{\columnwidth}{!}{%
    \begin{tabular}{l c c c}
        \toprule
        \textbf{Model class} & \textbf{Member Pass@1} & \textbf{Homogeneous} & \textbf{Heterogeneous} \\
        \midrule
        Consumer (Qwen3-8B $\times$3 vs.\ mid-tier) & 0.467 & 0.724 & \textbf{0.825} \\
        High-perf (Qwen3-Next-80B $\times$3 vs.\ high-end) & 0.700 & 0.889 & \textbf{0.911} \\
        \bottomrule
    \end{tabular}}
\end{table}

\definecolor{seriesA}{HTML}{2A78D6}
\definecolor{seriesB}{HTML}{EB6834}
\definecolor{seriesC}{HTML}{1BAF7A}
\begin{figure}[t]
    \centering
    \begin{tikzpicture}
        \begin{axis}[
                width=0.80\columnwidth, height=0.62\columnwidth,
                axis y line*=left, axis x line*=bottom, ylabel near ticks,
                xlabel={Threshold (\% of max.\ voting power)},
                ylabel={Accuracy change (pts), solid},
                xmin=0, xmax=100, ymin=-11, ymax=3,
                xtick={0,10,...,100}, ytick={-10,-8,-6,-4,-2,0,2},
                grid=major, grid style={line width=0.5pt, draw=black!15},
                legend style={font=\tiny, draw=none, at={(0.5,1.03)}, anchor=south, legend columns=3},
                label style={font=\footnotesize}, tick label style={font=\footnotesize},
                axis line style={line width=0.5pt}, tick style={line width=0.5pt}]
            \addplot[black!60, dashed, line width=0.5pt, forget plot] coordinates {(0,0) (100,0)};
            \addplot[black, densely dotted, line width=0.5pt, forget plot] coordinates {(90,-11) (90,3)};
            \addplot[black, densely dotted, line width=0.5pt, forget plot] coordinates {(70,-11) (70,3)};
            \addplot[color=seriesA, mark=*, mark size=1.6pt, line width=0.8pt]
                coordinates {(0,-10.0) (10,-9.2) (20,-8.3) (30,-5.8) (40,-5.0) (50,-2.5) (60,0.0) (70,-1.7) (80,0.0) (90,0.8) (100,0.0)};
            \addlegendentry{AIME mid-tier}
            \addplot[color=seriesB, mark=square*, mark size=1.6pt, line width=0.8pt]
                coordinates {(0,-7.8) (10,-7.8) (20,-7.8) (30,-6.7) (40,-6.7) (50,-5.6) (60,-2.2) (70,-1.1) (80,-1.1) (90,0.0) (100,0.0)};
            \addlegendentry{AIME high-end}
            \addplot[color=seriesC, mark=triangle*, mark size=1.9pt, line width=0.8pt]
                coordinates {(0,-9.0) (10,-9.0) (20,-8.7) (30,-8.7)(40,-8.1) (50,-6.4) (60,-2.3) (70,0.3) (80,0.9) (90,0.6) (100,0.0)};
            \addlegendentry{LCB v5 hard}
        \end{axis}
        \begin{axis}[
                width=0.80\columnwidth, height=0.62\columnwidth,
                axis y line*=right, axis x line=none, yticklabel pos=right, ylabel near ticks,
                ylabel={Tokens saved (\%), dashed},
                xmin=0, xmax=100, ymin=0, ymax=100, ytick={0,20,40,60,80,100},
                label style={font=\footnotesize}, tick label style={font=\footnotesize},
                axis line style={line width=0.5pt}, tick style={line width=0.5pt}]
            \addplot[color=seriesA, mark=o, mark size=1.6pt, line width=0.7pt, dashed, mark options={solid}]
                coordinates {(0,89) (10,89) (20,88) (30,87) (40,85) (50,83) (60,76) (70,69) (80,58) (90,47) (100,0)};
            \addplot[color=seriesB, mark=square, mark size=1.6pt, line width=0.7pt, dashed, mark options={solid}]
                coordinates {(0,89) (10,89) (20,89) (30,88) (40,88) (50,87) (60,85) (70,81) (80,77) (90,68) (100,0)};
            \addplot[color=seriesC, mark=triangle, mark size=1.9pt, line width=0.7pt, dashed, mark options={solid}]
                coordinates {(0,89) (10,89) (20,88) (30,88) (40,87) (50,85) (60,78) (70,61) (80,39) (90,19) (100,0)};
        \end{axis}
    \end{tikzpicture}
    \caption{Stopping threshold versus accuracy change relative to the full-length run (solid, left axis) and rounds saved (dashed, right axis). The dotted vertical line marks the 90\% (AIME) and 70\% (LiveCodeBench) threshold used in Table~\ref{tab:global_comparison}.}
    \label{fig:stopping_tradeoff}
\end{figure}
\begin{table}[ht]
    \centering
    \caption{Convergence based stopping ($\theta=0.9$ for AIME, $0.7$ for LiveCodeBench) compared to stopping at $T_{opt}$ or $T_{max}$}
    \label{tab:stopping_matrix}
    \small
    \begin{tabular}{l c c c}
        \toprule
        \textbf{Run} & \textbf{$T_{max}$} & \textbf{$T_{opt}$} & \textbf{Converged} \\
        \midrule
        AIME mid-tier & 79.2\% (7) & \textbf{82.5\%} (6) & 80.0\% (\textbf{3.8}) \\
        AIME high-end & 90.0\% (8) & \textbf{91.1\%} (5) & 90.0\% (\textbf{2.7}) \\
        LCB v5 hard   & 70.9\% (7) & \textbf{71.8\%} (5) & 71.2\% (\textbf{2.9}) \\
        \bottomrule
    \end{tabular}
\end{table}

\section{Discussion}
	\label{sec:discussion}

	While our implementation relies on discrete natural language strings ($y(t) \in \Sigma^*$) as the universal interface, this approach is inherently lossy. By sampling a single token at each step, the generator discards the rich probability distribution contained in the tail logits, resulting in pruning potentially valid reasoning paths before they can be evaluated.
	Instead of emitting a collapsed string, a future iteration of protocol could modify output harness to capture a \textbf{Token Lattice} $\mathcal{L}$ (we further discussed in Appendix \ref{app:lattices}).

\paragraph{Efficiency.} State-of-the-art MoEs face a specific bottleneck: while their \textit{compute} cost (FLOPs) is low due to sparse activation, their \textit{memory} cost is strictly bounded by the total parameter count. This requires high-bandwidth memory already caused severe market shortages. If compared to similar by benchmark performances MoE models at the time, we can see how MoM compares to standalone MoE such as GPT-OSS-120b, Qwen-3-235B, while high-end tier compares to proprietary SOTA like GPT-5. Storing 235B parameters at FP4 quantization requires $\approx 117.5 \text{ GB}$ of VRAM (plus KV cache overhead).  This exceeds the capacity of consumer high-end GPUs and requires enterprise systems and tensor parallelism handling. Comparable MoM ensemble contrary can run at consumer grade GPU cluster. Table \ref{tab:hardware_cost} details the Total Cost of Ownership (TCO) comparison, from a market price at the time this shows that our mid-tier can provide same accuracy results at \textbf{6-10x less capital expenditure cost}  (detailed experimental setup can be read in Appendix \ref{app:hwc}).

Time-inference wise our protocol is inference-time solution, if we treat a model (even CoT) as one-shot, then MoM requires several sequential rounds (on average 2.7-3.8 in our measurements) with convergence stopping signal in place.
Equation~\ref{eq:thermo_governing} also gives a deployment test: an ensemble improves only if evaluators grade more accurately than it generates ($p_v > p_g$), which can be measured on a small labelled calibration set.

\paragraph{Safety.} One of the most compelling features of the NSED protocol is its potential to improve end-user safety. The protocol exposes internal system metrics, such as the convergence or divergence of the ensemble's voting distribution. This provides a natural mechanism for confidence-based halting or human-in-the-loop (HITL) escalation.

Regulators increasingly demand that AI systems remain transparent and resilient to unforeseen failure modes, faults, and inconsistencies. Specifically, the EU AI Act \cite{eu_ai_act_2024} mandates that high-risk systems mitigate automation bias (Article 14), prevent recursive feedback loops (Article 15), and maintain mechanisms for bias correction and adversarial testing (Articles 10 and 55).

In this context, the DarkBench results which demonstrated a \textbf{37.5\% reduction in sycophancy and lower overall safety risk mean} (Table \ref{tab:darkbench}) represent a strong signal for regulatory compliance. By enforcing objective peer review over reputational weight via the diagonal mask, the protocol directly mitigates the "blind reliance" on single-model outputs warned against by regulators. Further safety studies, however, are necessary to explore the specific effects of injecting dedicated guardrail or adversarial models into the ensemble, as well as to map boundary failure modes.

\paragraph{User Experience.} Proposed system clearly transitions from chat-bot style to a high level objective solving, this aligns with trend in LLM implementations today that transitions to prolonged CoT workstyle, in that context we simply take it step further as distributed (system 3) thinking. That gives opportunity to rethink how humans should interact with Large Language Models. More precise, thorough inference is well suited to executive reporting style, while synchronous execution and timing planning allows implementing just-in-time task completion - regardless of system degradation or convergence, user has readable \textit{complete solution} that evolves and can be further propagated even before finalisation happened.

Because the protocol establishes clear timing constraints between synchronization phases and treats all communication channels as natural language, HITL oversight can be integrated directly into the semantic network architecture. This allows the system to seamlessly pause deliberation and escalate ambiguous cases to human experts ad-hoc, fulfilling strict regulatory demands for human agency and oversight.

\paragraph{Distributed system redundancy.} In regular neural network or MoE, a failure of one GPU will cause model to stop. Providers have to solve this with application level load balancers which implies resetting kv-cache even if only one GPU of a cluster malfunctions. contrary our solution allows hot swap modules, and system generally can be designed to be single-point failure free due to fact that orchestration kernel itself is a lightweight, deterministic program that can be solved by distributed nodes as BFT problem \cite{lamport1982byzantine} as well.

\subsection{Future work}
\paragraph{Continual learning.}
Proposed protocol is essentially high level Semantic Recursive Neural Network that does a forward-pass during generation. This intrinsically means that individual models and ensemble as joint cluster can train efficiently on it's data. Evaluation signals are directly usable for reinforcement learning between network clusters and even could be considered to be a privacy preserving way as models do not need to expose their data or private context to share evaluations.
Dynamic broker may learn higher level, architectural patterns of which models and which compositions work best together over time \textbf{without looking in to actual content}, all broker needs to see is evaluation scores.

By having direct reasoning deliberation traces, further work may introduce work on parameter-efficient updates (e.g., LoRA) to the base models using their own verified deliberation traces, we can achieve an \textbf{Autopoietic Improvement Cycle} \cite{zelikman2022star}. This mechanism mimics human experience replay where short-term "working memory" (deliberation) is encoded into long-term "weights" (synaptic plasticity) during idle periods.

\paragraph{NSED-Native Adapters.}
Although our ablation studies indicate that ensemble consisting of same model suffers from reduced variance, the logistical benefits of a unified substrate such as shared KV-caches and predictable latency remain compelling.

Future work might involve reconciling these conflicting constraints (Diversity vs. Uniformity) with use of role specific Low-Rank Adapters (LoRA) \cite{hu2021lora} or special embedding layers. Instead of deploying entirely distinct model architectures (e.g., Qwen vs. Llama), a single foundational model can be dynamically "colored" at runtime. A "Critic Adapter" can be active during the evaluation phase to maximize scrutiny, while a "Creative Adapter" activates during generation to maximize entropy.

This approach would allow decoupling general reasoning of base model from protocol roles of such adapters, enabling:
\begin{itemize}
	\item \textbf{Safety Adapters:} Specialized modules trained on datasets like DarkBench to rigorously audit proposals for manipulative patterns or bias, independent of the generator's alignment.
	\item \textbf{Domain Adapters:} Modules fine-tuned on niche corpora (e.g., Legal, Medical) to inject specific knowledge without retraining the verifying agents.
\end{itemize}
These lightweight modules could be dynamically swapped by the "Interface-Driven Constructor," creating a marketplace for modular cognitive skills while maintaining the \textbf{Computational Uniformity} required for efficient batching and caching.

Such shared substrate also permits denser protocol than text: agents could exchange logits or hidden states depending on networking bus constraints.

\paragraph{Composition and scale.} We tested single NSED layer, demonstrating Mixture-of-Models as system thinking approach. Open questions remain how accuracy and communication will scale when such layers are stacked or nested. Addressing this first requires adapting base models and training them to work with protocol context overhead. Future work should answer whether per-round sparsity will demonstrate scalability and further accuracy improvements. A broker module trained on convergence signals is a switch-transformer style router over multiple LLMs, it could also choose the graph shape and should be further studied.

\section{Conclusion}
The transition from static pre-training to dynamic inference-time compute and agentic, executive oriented workflow usage necessitates a fundamental re-evaluation of cognitive architectures. In this paper, we introduced the Mixture-of-Models approach to construct composable self-evaluating protocol. Reasoning is not solely a function of parameter count, but of topology and can be distributed. By formalizing the multi-agent interaction as a recurrent loop, we bridge the gap between connectionist control theory and agentic workflows.

With convergence-based stopping, an ensemble of three smaller models of at most 21B parameters outperforms every one of its members on two distinct benchmarks while no model is best at both. The ensemble improves on its average member by 48\% on AIME 2025 and 75\% on LiveCodeBench. This establishes a frontier for distributed intellect, showing that deliberative process can offset model scale. By trading time for recurrent refinement, we enable disjointed clusters of commodity hardware to reason together.
Ablation study shows that quadratic voting improves accuracy, and the convergence signal of the vote serves as a stopping indicator, suggesting that social-choice protocols such as the Delphi method and quadratic voting are usable building blocks for AI systems. While our current experiments emulated agent switching, the observed telemetry suggests the potential for a fully dynamic Broker that optimizes team composition based on historical signal-to-noise ratios or real-time signals. Looking forward, the integration of our proposed offline consolidation loops promises to evolve this architecture into a system that optimizes its energy expenditure in pursuit of semantic convergence. We posit that this shift from bigger weights to better circuits represents a necessary path toward sustainable Collective Intellect.

\bibliography{references}

\begin{thebibliography}{44}
\providecommand{\natexlab}[1]{#1}
\providecommand{\url}[1]{\texttt{#1}}
\expandafter\ifx\csname urlstyle\endcsname\relax
  \providecommand{\doi}[1]{doi: #1}\else
  \providecommand{\doi}{doi: \begingroup \urlstyle{rm}\Url}\fi

\bibitem[Behrouz et~al.(2024)Behrouz, Zhong, and Mirrokni]{behrouz2024titans}
Behrouz, A., Zhong, P., and Mirrokni, V.
\newblock Titans: Learning to memorize at test time, 2024.
\newblock URL \url{https://arxiv.org/abs/2501.00663}.

\bibitem[Behrouz et~al.(accessed at 3 jan 2026)Behrouz, Razaviyayn, Zhong, and
  Mirrokni]{behrouz2025nested}
Behrouz, A., Razaviyayn, M., Zhong, P., and Mirrokni, V.
\newblock Nested learning: The illusion of deep learning architecture, accessed
  at 3 jan 2026.
\newblock URL \url{https://abehrouz.github.io/files/NL.pdf}.

\bibitem[Besta et~al.(2024)Besta, Blach, Kubicek, Gerstenberger, Gianinazzi,
  Gajda, Lehmann, Podstawski, Niewiadomski, Nolle, and Hoefler]{besta2024graph}
Besta, M., Blach, N., Kubicek, A., Gerstenberger, R., Gianinazzi, L., Gajda,
  J., Lehmann, T., Podstawski, M., Niewiadomski, H., Nolle, P., and Hoefler, T.
\newblock Graph of thoughts: Solving elaborate problems with large language
  models.
\newblock In \emph{Proceedings of the AAAI Conference on Artificial
  Intelligence}, volume~38, pp.\  17682--17690, 2024.

\bibitem[Boizard et~al.(2024)]{boizard2024universal}
Boizard, N. et~al.
\newblock Towards cross-tokenizer distillation: the universal logit
  distillation loss for llms.
\newblock \emph{arXiv preprint arXiv:2402.12030}, 2024.

\bibitem[Breiman(2001)]{breiman2001rashomon}
Breiman, L.
\newblock Statistical modeling: The two cultures.
\newblock \emph{Statistical Science}, 16\penalty0 (3):\penalty0 199--231, 2001.

\bibitem[Brown et~al.(2024)Brown, Juravsky, Ehrlich, Clark, Le, Ré, and
  Mirhoseini]{brown2024monkeys}
Brown, B., Juravsky, J., Ehrlich, R., Clark, R., Le, Q.~V., Ré, C., and
  Mirhoseini, A.
\newblock Large language monkeys: Scaling inference compute with repeated
  sampling, 2024.
\newblock URL \url{https://arxiv.org/abs/2407.21787}.

\bibitem[Cemri et~al.(2025)Cemri, Pan, Yang, Agrawal, Chopra, Tiwari, Keutzer,
  Parameswaran, Klein, Ramchandran, Zaharia, Gonzalez, and
  Stoica]{cemri2025multiagentllmsystemsfail}
Cemri, M., Pan, M.~Z., Yang, S., Agrawal, L.~A., Chopra, B., Tiwari, R.,
  Keutzer, K., Parameswaran, A., Klein, D., Ramchandran, K., Zaharia, M.,
  Gonzalez, J.~E., and Stoica, I.
\newblock Why do multi-agent llm systems fail?, 2025.
\newblock URL \url{https://arxiv.org/abs/2503.13657}.

\bibitem[Cobbe et~al.(2021)Cobbe, Kosaraju, Bavarian, Chen, Jun, Kaiser,
  Plappert, Tworek, Hilton, Nakano, Hesse, and Schulman]{cobbe2021training}
Cobbe, K., Kosaraju, V., Bavarian, M., Chen, M., Jun, H., Kaiser, L., Plappert,
  M., Tworek, J., Hilton, J., Nakano, R., Hesse, C., and Schulman, J.
\newblock Training verifiers to solve math word problems.
\newblock \emph{arXiv preprint arXiv:2110.14168}, 2021.
\newblock Empirically demonstrates that a dedicated verifier scales more
  effectively with data than a generator, establishing the existence of a
  'Verification Gap' where $p_{verify} > p_{generate}$.

\bibitem[Cover(1965)]{cover1965}
Cover, T.~M.
\newblock Geometrical and statistical properties of systems of linear
  inequalities with applications in pattern recognition.
\newblock \emph{IEEE transactions on electronic computers}, \penalty0
  (3):\penalty0 326--334, 1965.

\bibitem[Cuturi(2013)]{cuturi2013sinkhorn}
Cuturi, M.
\newblock Sinkhorn distances: Lightspeed computation of optimal transport.
\newblock In \emph{Advances in Neural Information Processing Systems},
  volume~26, pp.\  2292--2300, 2013.

\bibitem[de~Condorcet(1785)]{condorcet1785essai}
de~Condorcet, M.
\newblock \emph{Essai sur l’application de l’analyse à la probabilité des
  décisions rendues à la pluralité des voix}.
\newblock Imprimerie Royale, Paris, 1785.
\newblock Establishes that if individual voters are better than random
  ($p>0.5$), the probability of a correct majority vote approaches 1 as the
  number of voters increases.

\bibitem[DeepSeek-AI et~al.(2025)DeepSeek-AI, Guo, and et. al]{deepseekr1}
DeepSeek-AI, Guo, D., and et. al.
\newblock Deepseek-r1: Incentivizing reasoning capability in llms via
  reinforcement learning, 2025.
\newblock URL \url{https://arxiv.org/abs/2501.12948}.

\bibitem[Eigen et~al.(2014)Eigen, Ranzato, and
  Sutskever]{eigen2014learningfactoredrepresentationsdeep}
Eigen, D., Ranzato, M., and Sutskever, I.
\newblock Learning factored representations in a deep mixture of experts.
\newblock 2014.
\newblock URL \url{https://arxiv.org/abs/1312.4314}.

\bibitem[{European Parliament and Council of the European
  Union}(2024)]{eu_ai_act_2024}
{European Parliament and Council of the European Union}.
\newblock Regulation ({EU}) 2024/1689 of the european parliament and of the
  council of 13 june 2024 laying down harmonised rules on artificial
  intelligence ({Artificial Intelligence Act}).
\newblock \emph{Official Journal of the European Union}, L 1689:\penalty0 1,
  2024.
\newblock URL
  \url{https://eur-lex.europa.eu/legal-content/EN/TXT/HTML/?uri=CELEX:02024R1689-20260727}.
\newblock Consolidated text including amendments up to Regulation (EU)
  2026/1744.

\bibitem[Fedus et~al.(2022)Fedus, Zoph, and
  Shazeer]{fedus2022switchtransformersscalingtrillion}
Fedus, W., Zoph, B., and Shazeer, N.
\newblock Switch transformers: Scaling to trillion parameter models with simple
  and efficient sparsity, 2022.
\newblock URL \url{https://arxiv.org/abs/2101.03961}.

\bibitem[Graves(2016)]{graves2016adaptive}
Graves, A.
\newblock Adaptive computation time for recurrent neural networks.
\newblock \emph{arXiv preprint arXiv:1603.08983}, 2016.

\bibitem[Hu et~al.(2022)Hu, Shen, Wallis, Allen-Zhu, Li, Wang, Wang, and
  Chen]{hu2021lora}
Hu, E.~J., Shen, Y., Wallis, P., Allen-Zhu, Z., Li, Y., Wang, S., Wang, L., and
  Chen, W.
\newblock Lora: Low-rank adaptation of large language models.
\newblock In \emph{International Conference on Learning Representations}, 2022.
\newblock URL \url{https://openreview.net/forum?id=nZeVKeeFYf9}.

\bibitem[Jacobs et~al.(1991)Jacobs, Jordan, Nowlan, and Hinton]{Jacobs91}
Jacobs, R., Jordan, M., Nowlan, S., and Hinton, G.
\newblock Adaptive mixtures of local experts.
\newblock \emph{Neural Computation}, 3:\penalty0 79--87, 03 1991.
\newblock \doi{10.1162/neco.1991.3.1.79}.

\bibitem[Jaech \& et. al(2024)Jaech and et. al]{openai2024openaio1card}
Jaech, A. and et. al.
\newblock Openai o1 system card, 2024.
\newblock URL \url{https://arxiv.org/abs/2412.16720}.

\bibitem[Jin et~al.(2024)Jin, Zhu, Yuan, and Yan]{jin2024moeplus}
Jin, P., Zhu, B., Yuan, L., and Yan, S.
\newblock Moe++: Accelerating mixture-of-experts methods with zero-computation
  experts, 2024.
\newblock URL \url{https://arxiv.org/abs/2410.07348}.

\bibitem[Kran et~al.(2025)Kran, Nguyen, Kundu, Jawhar, Park, and
  Jurewicz]{kran2025darkbenchbenchmarkingdarkpatterns}
Kran, E., Nguyen, H. M.~J., Kundu, A., Jawhar, S., Park, J., and Jurewicz,
  M.~M.
\newblock Darkbench: Benchmarking dark patterns in large language models, 2025.
\newblock URL \url{https://arxiv.org/abs/2503.10728}.

\bibitem[Lalley \& Weyl(2018)Lalley and Weyl]{lalley2018quadratic}
Lalley, S. and Weyl, E.
\newblock Quadratic voting: How mechanism design can radicalize democracy,
  2018.
\newblock URL \url{http://dx.doi.org/10.2139/ssrn.2003531}.

\bibitem[Lamport et~al.(1982)Lamport, Shostak, and Pease]{lamport1982byzantine}
Lamport, L., Shostak, R., and Pease, M.
\newblock The byzantine generals problem.
\newblock \emph{ACM Transactions on Programming Languages and Systems
  (TOPLAS)}, 4\penalty0 (3):\penalty0 382--401, 1982.
\newblock \doi{10.1145/357172.357176}.

\bibitem[Lightman et~al.(2023)Lightman, Kosaraju, Burda, Edwards, Baker, Lee,
  Leike, Schulman, Sutskever, and Cobbe]{lightman2023lets}
Lightman, H., Kosaraju, V., Burda, Y., Edwards, H., Baker, B., Lee, T., Leike,
  J., Schulman, J., Sutskever, I., and Cobbe, K.
\newblock Let's verify step by step.
\newblock \emph{arXiv preprint arXiv:2305.20050}, 2023.
\newblock Shows that Process Reward Models (PRMs) significantly outperform
  outcome-based supervision, validating the NSED approach of step-wise
  consensus.

\bibitem[Liu et~al.(2023)Liu, Lin, Hewitt, Paranjape, Bevilacqua, Petroni, and
  Liang]{liu2023lostmiddlelanguagemodels}
Liu, N.~F., Lin, K., Hewitt, J., Paranjape, A., Bevilacqua, M., Petroni, F.,
  and Liang, P.
\newblock Lost in the middle: How language models use long contexts, 2023.
\newblock URL \url{https://arxiv.org/abs/2307.03172}.

\bibitem[Minixhofer et~al.(2025)Minixhofer, Vulić, and
  Ponti]{minixhofer2025universal}
Minixhofer, B., Vulić, I., and Ponti, E.~M.
\newblock Universal cross-tokenizer distillation via approximate likelihood
  matching, 2025.
\newblock URL \url{https://arxiv.org/abs/2503.20083}.

\bibitem[Peyr{\'e} \& Cuturi(2019)Peyr{\'e} and Cuturi]{peyre2019computational}
Peyr{\'e}, G. and Cuturi, M.
\newblock Computational optimal transport.
\newblock \emph{Foundations and Trends in Machine Learning}, 11\penalty0
  (5-6):\penalty0 355--607, 2019.

\bibitem[Puchinger et~al.()Puchinger, Raidl, and Pferschy]{jakob2015}
Puchinger, J., Raidl, G.~R., and Pferschy, U.
\newblock The multidimensional knapsack problem: Structure and algorithms.
\newblock URL \url{https://inria.hal.science/hal-01224914v1/document}.

\bibitem[Rescher(1998)]{rescher1998predicting}
Rescher, N.
\newblock \emph{Predicting the future: An introduction to the theory of
  forecasting}.
\newblock State University of New York Press, Albany, 1998.
\newblock ISBN 978-0-7914-3553-3.

\bibitem[Schwartz et~al.(2020)Schwartz, Dodge, Smith, and
  Etzioni]{schwartz2020green}
Schwartz, R., Dodge, J., Smith, N.~A., and Etzioni, O.
\newblock Green ai.
\newblock \emph{Communications of the ACM}, 63\penalty0 (12):\penalty0 54--63,
  2020.

\bibitem[Sharma et~al.(2024)Sharma, Tong, Korbak, Duvenaud, Askell, Bowman,
  Cheng, Durmus, Hatfield-Dodds, Johnston, Kravec, Maxwell, McCandlish,
  Ndousse, Rausch, Schiefer, Yan, Zhang, and Perez]{sharma2024understanding}
Sharma, M., Tong, M., Korbak, T., Duvenaud, D., Askell, A., Bowman, S.~R.,
  Cheng, N., Durmus, E., Hatfield-Dodds, Z., Johnston, S.~R., Kravec, S.,
  Maxwell, T., McCandlish, S., Ndousse, K., Rausch, O., Schiefer, N., Yan, D.,
  Zhang, M., and Perez, E.
\newblock Towards understanding sycophancy in language models, 2024.
\newblock URL \url{https://arxiv.org/abs/2310.13548}.

\bibitem[Shumailov et~al.(2023)Shumailov, Shumaylov, Zhao, Gal, Papernot, and
  Anderson]{shumailov2023curse}
Shumailov, I., Shumaylov, Z., Zhao, Y., Gal, Y., Papernot, N., and Anderson, R.
\newblock The curse of recursion: Training on generated data makes models
  forget.
\newblock \emph{arXiv preprint arXiv:2305.17493}, 2023.
\newblock Also published in Nature (2024) as 'AI models collapse when trained
  on recursively generated data'.

\bibitem[Snell et~al.(2024)Snell, Lee, Xu, and Kumar]{snell2024scaling}
Snell, C., Lee, J., Xu, K., and Kumar, A.
\newblock Scaling llm test-time compute optimally can be more effective than
  scaling model parameters, 2024.
\newblock URL \url{https://arxiv.org/abs/2408.03314}.

\bibitem[Tandon \& et. al(2025)Tandon and et. al]{tandon2025ttt}
Tandon, A. and et. al.
\newblock End-to-end test-time training for long context, 2025.
\newblock URL \url{https://arxiv.org/abs/2512.23675}.

\bibitem[Vaswani et~al.(2023)Vaswani, Shazeer, Parmar, Uszkoreit, Jones, Gomez,
  Kaiser, and Polosukhin]{vaswani2023attentionneed}
Vaswani, A., Shazeer, N., Parmar, N., Uszkoreit, J., Jones, L., Gomez, A.~N.,
  Kaiser, L., and Polosukhin, I.
\newblock Attention is all you need, 2023.
\newblock URL \url{https://arxiv.org/abs/1706.03762}.

\bibitem[Wald(1945)]{wald1945}
Wald, A.
\newblock \emph{Sequential Tests of Statistical Hypotheses}.
\newblock John Wiley \& Sons, 1945.

\bibitem[Wang et~al.(2024)Wang, Wang, Athiwaratkun, Zhang, and
  Zou]{wang2024mixture}
Wang, J., Wang, J., Athiwaratkun, B., Zhang, C., and Zou, J.
\newblock Mixture-of-agents enhances large language model capabilities, 2024.
\newblock URL \url{https://arxiv.org/abs/2406.04692}.

\bibitem[Wu et~al.(2023)Wu, Bansal, Zhang, Wu, Li, Zhu, Jiang, Zhang, Zhang,
  Liu, Awadallah, White, Burger, and Wang]{AutoGen23}
Wu, Q., Bansal, G., Zhang, J., Wu, Y., Li, B., Zhu, E., Jiang, L., Zhang, X.,
  Zhang, S., Liu, J., Awadallah, A.~H., White, R.~W., Burger, D., and Wang, C.
\newblock Autogen: Enabling next-gen llm applications via multi-agent
  conversation.
\newblock \emph{arXiv preprint arXiv:2308.08155}, 2023.
\newblock URL \url{https://arxiv.org/abs/2308.08155}.

\bibitem[Yao et~al.(2023)Yao, Zhao, Yu, Du, Shafran, Narasimhan, and
  Cao]{yao2023reactsynergizingreasoningacting}
Yao, S., Zhao, J., Yu, D., Du, N., Shafran, I., Narasimhan, K., and Cao, Y.
\newblock React: Synergizing reasoning and acting in language models, 2023.
\newblock URL \url{https://arxiv.org/abs/2210.03629}.

\bibitem[Yao et~al.(2024)Yao, Yu, Zhao, Shafran, Griffiths, Cao, and
  Narasimhan]{yao2024tree}
Yao, S., Yu, D., Zhao, J., Shafran, I., Griffiths, T.~L., Cao, Y., and
  Narasimhan, K.
\newblock Tree of thoughts: Deliberate problem solving with large language
  models.
\newblock In \emph{Advances in Neural Information Processing Systems},
  volume~36, pp.\  11830--11843, 2024.

\bibitem[Yi et~al.(2025)Yi, Zhou, Cao, Niu, Liu, and
  Han]{yi2025debateequilibriumbeliefdrivenmultiagent}
Yi, X., Zhou, Z., Cao, C., Niu, Q., Liu, T., and Han, B.
\newblock From debate to equilibrium: Belief-driven multi-agent llm reasoning
  via bayesian nash equilibrium, 2025.
\newblock URL \url{https://arxiv.org/abs/2506.08292}.

\bibitem[Zalta \& Nodelman()Zalta and Nodelman]{sfuEnc}
Zalta, C. P. E. E.~N. and Nodelman, U. (eds.).
\newblock \emph{The Stanford Encyclopedia of Philosophy}.
\newblock URL
  \url{https://plato.stanford.edu/entries/jury-theorems/#CondJuryTheo}.

\bibitem[Zelikman et~al.(2022)Zelikman, Wu, Mu, and Goodman]{zelikman2022star}
Zelikman, E., Wu, Y., Mu, J., and Goodman, N.~D.
\newblock Star: Bootstrapping reasoning with reasoning.
\newblock 2022.
\newblock URL \url{https://arxiv.org/abs/2203.14465}.

\bibitem[Zhang et~al.(2024)Zhang, Sun, Chen, Pfister, Zhang, and
  Arik]{zhang2024chain}
Zhang, Y., Sun, R., Chen, Y., Pfister, T., Zhang, R., and Arik, S.~{\"O}.
\newblock Chain of agents: Large language models collaborating on long-context
  tasks, 2024.
\newblock URL \url{https://arxiv.org/abs/2406.02818}.

\end{thebibliography}
\bibliographystyle{mlsys2025}

\appendix
\section{AI Usage Disclosure}
Generative AI was used in this work during ideation stage, literature research, writing code and partially used in generating agent prompts. We used LLMs to proof-read and check our paper LaTeX against inconsistencies in our data and algorithms and as a pre-submission reviewer.
\section{Derivation of the Governing Equation}
\label{app:derivation}
We derive this formulation based on Wald's Sequential Probability Ratio Test (SPRT), as defined by \citet{wald1945}. This gives rise to our gain term $1 - e^{-\Lambda (p_v - p_g)(t-1)}$, where $\Lambda$ represents the topological efficiency of the evidence integration.

Second, the driving force of this gain is determined by \textbf{Condorcet's Jury Theorem} \cite{sfuEnc, condorcet1785essai}. Convergence is only mathematically possible if the effective verification capability ($p_v$) exceeds the generation capability ($p_g$). We therefore scale the exponent by the "Signal Gap" $(p_v - p_g)$, representing the information gain per round.
\begin{table}[ht]
    \centering
    \caption{Eq.~\ref{eq:thermo_governing} parameters: $p_g$ (round-1 accuracy) and $p_v$ (mean evaluator grading accuracy) measured, $\Lambda$ and $\beta$ fitted. The gap $p_v - p_g$ is small (0.01--0.04), so $\Lambda$ is sensitive to how $p_v$ is measured.}
    \label{tab:fit}
    \small
    \setlength{\tabcolsep}{4pt}
    \begin{tabular}{l c c c c c c}
        \toprule
        \textbf{Run} & $p_g$ & $p_v$ & $\Lambda$ & $\beta$ & $R^2$ & $t_{opt}$ \\
        \midrule
        AIME mid  & 0.692 & 0.733 & 4.95 & 0.0030 & 0.98 & 5.3 \\
        AIME high & 0.822 & 0.852 & 8.65 & 0.0016 & 0.81 & 5.5 \\
        LCB       & 0.619 & 0.633 & 10.2 & 0.0039 & 0.84 & 4.9 \\
        \bottomrule
    \end{tabular}
\end{table}

For convergence-based stopping (Eq.~\ref{eq:weighted_convergence}) we fit $A_{norm}$ once, on AIME mid-tier, and use this one curve for all runs. Fitting it per run instead (2-fold cross-validated over questions) changes converged accuracy by at most 0.7 points, as the normalised curves of all three runs peak at round 5 and differ little in shape.

The quadratic penalty term $\beta t^2$ accounts for the \textbf{Entropic Fatigue} inherent in recursive semantic processes. This formulation is grounded in the $O(n^2)$ self-attention complexity of the Transformer architecture \cite{vaswani2023attentionneed}. As the deliberation history $\mathcal{H}$ expands, the signal-to-noise ratio degrades because the semantic "surface area" for potential hallucinations and sycophantic loops grows quadratically with the total token count. Unlike ideal Bayesian updates, LLMs suffer from context pollution, where the accumulation of failed reasoning traces and peer-disagreements eventually has a repulsive potential that overwhelms the marginal information gain of further rounds.

\section{System Control Flow \& The General Optimization Framework:}
At the control plane (Fig. \ref{fig:control_flow_agent_representation}), the system relies on the Dynamic Broker. Unlike static MoE routers that make simple per-token decisions, the Broker functions as a \textit{Session Constructor} and resource allocator. It is designed as an extensible optimization engine capable of ingesting diverse signals—from static benchmarks to real-time telemetry—to solve for the optimal team composition.
\section{Broker Optimisation Framework}\label{app:opt_fw} The Broker's core responsibility is to select a subset of agents $A \subset \mathcal{M}$ and an optimal deliberation budget $T$ (depth). This is formulated as a variation of the Multi-Dimensional Knapsack Problem \cite{jakob2015}, maximizing the thermodynamic utility function (Eq. \ref{eq:thermo_governing}) subject to the task's specifics.

The optimization objective is to solve for the tuple $\{A, T\}$:
\[
	\operatorname*{maximize}_{A, T} \quad E(A, T) = \text{Utility}(A, T | \mathcal{T}) - \lambda \cdot \text{Cost}(A, T)
\]
subject to:
\[
\left\{
\begin{aligned}
	\text{Latency}(A, T) & \le T_{max}^{SLA}     \\
	\text{Cost}(A, T)    & \le C_{max}^{SLA}     \\
	\text{Quality}(A)    & \ge \mathcal{Q}_{min}
\end{aligned}
\right.
\]

Where:
\begin{itemize}
	\item $\text{Utility}(A, T | \mathcal{T})$ is the predicted accuracy derived from historical influence matrices (see Section \ref{sec:broker_empiric}).
	\item $\text{Cost}(A, T) \approx \sum_{a \in A} (\text{Price}_a \times \text{Tokens}_a \times T)$.
	\item $\lambda \ge 0$ is a user-configurable elasticity coefficient.
\end{itemize}

This formulation ensures the Broker does not merely select the "smartest" agents, but identifies the specific \textbf{Hardware-Time configuration} that targets the peak of the thermodynamic curve ($T_{opt}$) before entropic fatigue sets in. The solver logic is implementation-agnostic, supporting various backends ranging from Constraint Satisfaction Solvers (CSP) to lightweight neural policy networks that map task embeddings to expert domains.

Architecturally, as depicted in Fig. \ref{fig:control_flow_agent_representation}, the Client interacts with the Orchestrator, which treats the Broker as an upstream "Intelligence Supplier," transparently fulfilling the request within the negotiated constraints.

\paragraph{The Session Manifest:}
The output of the optimization is not merely a list of agents, but a complete \textbf{Session Manifest} $\mathcal{M}_{session}$. Since the cost function is time-dependent, the Broker must solve for the \textbf{Optimal Control Policy} alongside the team composition.

The Broker returns a tuple $\mathcal{M}_{session} = \{ \mathcal{A}^*, \gamma(t) \}$, where:
\begin{itemize}
	\item $\mathcal{A}^*$: The optimal subset of expert agents to instantiate.
	\item $\gamma(t)$: The \textbf{Time-Variant Decay Policy}. Instead of a static scalar, $\gamma$ is defined as a function of the internal step counter $\gamma(t; T_{opt})$. This function governs the system's thermodynamic lifecycle:
	      \begin{equation}
		      \gamma(t) = \gamma_{base} \cdot \mathbb{I}(t < T_{opt})
	      \end{equation}
	      This clamps the feedback gain to zero when the fatigue limit is reached ($t \ge T_{opt}$). The neural circuit is programmed to terminate execution automatically when the recurrence signal $\gamma(t) \to 0$, ensuring the system never accumulates costs beyond the point of positive marginal utility.
\end{itemize}

This abstraction unifies the "Soft Horizon" (forgetting) and "Hard Horizon" (stopping) into a single governing function, simplifying the runtime logic into a pure feedback control loop.
\paragraph{Orchestrator role:} orchestrator acts as a central coordinator, managing the state of the execution and short-term memory specific to agent ReAct \cite{yao2023reactsynergizingreasoningacting} loops.

Since this is a synchronous barrier protocol, round latency is bounded by the slowest agent ($L_{round} = \max(t_i)$). To mitigate the "Straggler Problem," the Orchestrator maintains a real-time telemetry of the execution circuit. If an agent's response time exceeds thresholds, the Orchestrator may trigger a Hot-Swap Event, seamlessly replacing the stalled model with a reserve agent of equivalent capability to maintain the SLA.

We assume that Broker is a stateful component that maintains a historic track of record, receiving feedback on the agent performance from client and orchestrator such that allows refining and adapting the selection strategies in future.

This allows system to continuously learn and reinforce strategies as answer to ever changing conditions without explicit need to re-train the agents who are treated as off-shelf components.

\subsection{Anonymization}
The routing layer replaces model identities with persona names before sending each candidate to peer evaluators, and does not send an agent its own proposal.

\section{Detailed explanation of The Neural Circuit Representation}
At the data plane, we model the execution loop not as a conversation, but as a \textbf{Macro-Scale Semantic Recurrent Neural Network} (Fig.\ref{fig:neural_topology}). This is topologically isomorphic to a Long Short-Term Memory (LSTM) unit, where $N$ agents function as parallel processing gates that regulate the read/write operations to a global, mutable cell state (consensus).
\paragraph{Semantic vs. Gradient Recurrence:}
Unlike traditional RNNs which utilize Backpropagation Through Time (BPTT) to update floating-point weights, NSED uses \textbf{Semantic Feedback Loops}. The ``error signal'' is quantified as the divergence in the Quadratic Voting matrix and is not propagated via gradients, but is symbolically encoded into the Consensus State $S_t$. This state loops back to the input gate, modifying the effective attention of the frozen agents in the subsequent timestep.

\paragraph{Meta-Learning (Broker Feedback):}
The system allows a secondary feedback loop at the orchestration layer. Telemetry from the voting phase is ``backpropagated'' to the Dynamic Expertise Broker as a reinforcement signal, updating the selection weights for future sessions (as detailed in Section \ref{sec:broker_empiric}).

\section{Topological Governance and Attention Constraints}
\label{app:topology_blinding}

Unlike standard Multi-Agent Systems where agents communicate via unrestricted open-book dialogue, NSED enforces a strict trustless topology. The system architecture imposes hard constraints on the information flow to guarantee incentive compatibility and mitigate the ``Authority Bias'' often observed in heterogeneous ensembles.

\subsection{Blind Routing}
Sharma et.al \cite{sharma2024understanding} have demonstrated that LLMs are subject to authority biases when weaker model treats information about another reasoning source as evaluation basis. By blinding model names we prevent this risk as well as game theoretic possibilities that extended general intelligence systems might bribe each other. This architectural constraint forces the ensemble to converge on \textit{merit} rather than \textit{reputational weight}, simulating a double-blind peer review process at the protocol level.

\subsubsection{Historical Mapping}
While current proposals are blinded, the system exposes the \textbf{Historical Voting Matrix} ($V_{t-1}$) to all agents as a form of Recurrent State. This allows agents to map their attention based on \textit{social signal variance} rather than content alone.

If we define the controversy score of a previous proposal $k$ as the variance of its votes $\sigma^2(V_{:,k})$, agents are prompted to allocate higher cognitive resources to proposals where $\sigma^2 > \delta$ (high disagreement), and lower resources where $\sigma^2 \approx 0$ (consensus). This functions as a \textbf{Social Attention Mechanism}, optimizing the Token Budget ($C_{max}$) by focusing deliberation on unresolved edges of the graph.
\subsection{Extended The Agentic Oracle}\label{sec:sysarch_oracle}
While the core NSED protocol focuses on semantic verification, the architecture supports an \textbf{Agentic Oracle} (Fig. \ref{fig:control_flow_agent_representation}) extension. In this configuration, the Orchestrator provisions a dedicated, read-only side-channel to the environment (e.g., a file system or retrieval index) for each agent.

\paragraph{1. Parallel Context Widening:}
Instead of serializing a massive context into a single model's window, the Broker shards the retrieval task. Agents execute local ReAct loops to ingest distinct data shards via their side-channels.
\[ \text{Context}(A_{total}) = \bigcup_{i=1}^{N} \text{ReAct}(Agent_i, \text{Shard}_i) \]

\paragraph{2. The Compression Mechanism:}
The NSED deliberation loop functions as a \textbf{Semantic Zipper}. When Agent A (who read File X) proposes a solution, and Agent B (who read File Y) critiques it, the resulting consensus $y(t)$ represents the intersection of truths from both contexts. This allows the system to reason over datasets larger than any single agent's context window ($L_{sys} \gg L_{model}$).

\section{Detailed experimental setup}\label{app:setup}
To avoid complexity of logit translation and independent model vector normalisation we used natural language as composability layer, which is natural choice also for explainability and traceability of the system. Thus, we do think that ultimate goal is to actually allow uncertainties be translated via logit layers rather than natural language directly.
\paragraph{Core Process Elements:}
The deliberation process involves three primary entities:
\begin{enumerate}
	\item \textbf{Input Task ($Q$):} The initial prompt or problem statement that initiates the NSED barrier loop.
	\item \textbf{Expert Pool ($\mathcal{A}$):} The set of $N$ heterogeneous agents performing generation and evaluation.
	\item \textbf{Deliberation Rounds ($T$):} The synchronous execution steps. Each round consists of parallel output generation, N-to-N peer review, and feedback integration.
\end{enumerate}

\paragraph{Broker Configuration:}
For this experimental validation, the Dynamic Expertise Broker was configured with fixed, pre-determined agent profiles broken in two ensembles, providing simplest way of solving the Knapsack optimization at runtime for every prompt. We utilize the post-run telemetry to demonstrate how influence data can be fed back into the Broker to optimize future team composition in a production environment.
\subsection{Quadratic Voting Implementation}
Each evaluator scores every peer proposal with a raw vote $v_{j,i} \in [0, 100]$, and the budget is $B = 100$ (Eqs.~\ref{eq:self_mask} - \ref{eq:qv_activation}). A vote vector whose total exceeds $B$ is scaled down to it; otherwise $\hat{v}_{j,i} = v_{j,i}/100$. The square root then dampens the impact of any attempt to spend the whole budget on one option, while every QV score is in $[0, 1]$. In the implementation this is $\sqrt{v'_{j,i}}/10$, with $v'_{j,i}$ the budget-scaled vote.

\subsection{Consensus Selection Strategies}\label{app:setup:cons_strat}
To determine the optimal termination logic, we evaluated three distinct aggregation strategies against the ground truth:

\begin{enumerate}
	\item \textbf{NSED Consensus (Live):} Selects the highest-scoring proposal in the final round $T$. This assumes the latest state is the most refined.
	      \[ S_{live} = \max_{p \in R_T} (QV(p)) \]

	\item \textbf{Global History Max (Unweighted):} Selects the highest-scoring proposal observed across \textit{all} rounds ($1 \dots T$). This strategy protects against regression, where a correct early answer is abandoned.
	      \[ S_{hist} = \max_{p \in \mathcal{H}} (QV(p)) \]

	\item \textbf{Time-Weighted History:} Biases selection toward later rounds to reward convergence while retaining high-confidence history. We tested two weight kernels:
	      \begin{itemize}
		      \item \textit{Linear:} $\omega(t) = 1 + \eta (t-1)$ (tested with $\eta=0.05, 0.20$).
		     \item \textit{Exponential:} $\omega(t) = \kappa^{t-1}$ (tested with $\kappa=1.1$).
	      \end{itemize}
\end{enumerate}

\subsection{Prompt engineering}
We engineered prompts to maximize the protocol's heterogenous setup, we also observed during benchmarking that agents trend to gravitate towards false positive responses (rating high hallucinations), rather false negatives (rating low correct answers). To mitigate this, we prompted agents to be skeptical to avoid overconfidence.
Persona texts and full prompt templates are given in Appendix~\ref{app:prompts}

\paragraph{"Harsh" Scoring Directive: }
To counteract the premature convergence observed in preliminary runs—where agents blindly assigned high scores to peers—we injected a "Harsh Scoring" constraint into the evaluation prompt. Agents were explicitly instructed to use the full 0-100 continuous scale and verify every step rigorously before assigning credit. Table~\ref{tab:agent_stats} reports how often each evaluator graded a peer correctly and how often it endorsed an incorrect candidate.
\begin{table}[ht]
    \centering
    \caption{Evaluator statistics on AIME 2025.}
    \label{tab:agent_stats}
    \small
    \setlength{\tabcolsep}{4pt}
    \begin{tabular}{l l c c c}
        \toprule
        & \textbf{Evaluator} & \textbf{Acc.} & \textbf{False end.} & \textbf{$N$} \\
        \midrule
        Mid-tier & GPT-OSS-20B   & 0.82 & 13.2\% & 1514 \\
                 & Qwen3-8B      & 0.76 & 27.4\% & 1574 \\
                 & Gemma-3-12B   & 0.62 & 48.6\% & 1575 \\
        \midrule
        High-end & Qwen3-Next-80B & 0.92 & 38.0\% & 1373 \\
                 & GPT-OSS-120B   & 0.91 & 36.2\% & 1176 \\
                 & DeepSeek-V3.1  & 0.74 & 38.8\% & 1399 \\  %
        \bottomrule
    \end{tabular}
\end{table}

\subsection{Semantic Implementation of Gamma}
Consistent with our principle of semantic isomorphism, the attention decay factor $\gamma$ was implemented via Context Windowing rather than vector scaling.
\begin{itemize}
	\item \textbf{Sliding History:} We utilized a persistent state container. As the round count $t$ increases, proposals from rounds before $t-1$ are displaced from the "Active Working Memory" object to a "Historical" block with older data compressed in to a ReAct tool call \texttt{read\_proposal(round,agent\_id)}. This lowers the attention weight ($\gamma$) on older data, forcing agents to focus on recent developments.
	\item \textbf{Hard Stop:} The termination condition was strictly enforced via a loop counter, with $T_{max} = 7$ for standard runs and $T_{max} = 8$ for high-performance runs, preventing infinite recursive loops.
\end{itemize}
\subsection{Tool Calling \& Action Space (The Protocol API)}
Unlike standard "Chain-of-Thought" benchmarks where models output free text, NSED agents interact with the environment strictly via a defined Function Calling API. This enforces structural integrity and allows the Orchestrator to parse reasoning traces deterministically.

\begin{itemize}
	\item \textbf{The Action Space:} Agents are provided with two primary protocol tools:
	      \begin{itemize}
		      \item \texttt{submit\_proposal(thought\_process, solution\_content)}: ends the proposing phase. For code tasks, \texttt{solution\_content} holds the full source.
		      \item \texttt{submit\_batch\_\allowbreak{}evaluation(evaluations)}: ends the evaluating phase with one entry per peer (\texttt{agent\_id}, \texttt{endorsement\_weight} in $[0, 100]$, \texttt{justification}, \texttt{is\_final\_solution}).
	      \end{itemize}
	\item \textbf{Scratchpad:} In order to save agent context memory and compress findings, we provided agents with \texttt{update\_scratchpad(content, mode='append' | 'overwrite')} to save summarized findings of previous rounds and ReAct loop calls; earlier rounds can be read in full with \texttt{read\_proposal}, \texttt{read\_critiques} and \texttt{read\_own\_proposal}.
	\item \textbf{Dual-Mode Parsing Strategy:} Given the heterogeneous nature of the ensemble, we observed that smaller open-weight models often suffer from "Tool Hallucination" (writing the JSON payload as text rather than emitting the special control tokens). To mitigate this, we implemented a Middleware Interceptor:
	      \begin{itemize}
		      \item \textit{Native Execution:} For high-compliance models (e.g., Qwen 3), we utilized the native \texttt{nous} or tool-use formats.
		      \item \textit{Heuristic Unwrapping:} The middleware that uses regex heuristics to detect and execute "pseudo-calls" in the output stream. This ensured that "Format Errors" did not falsely penalize semantic intelligence.
	      \end{itemize}
\end{itemize}
\subsection{Hyperparameter Configuration.}
All agents use a presence penalty of $\alpha = 1.5$. Preliminary grid searches indicated that standard values ($\alpha \approx 0.0$) led to immediate "consensus collapse" in Code Generation tasks, while excessive penalties ($\alpha > 2.0$) caused semantic drift. The value $\alpha=1.5$ was empirically determined to provide balanced structural coherence with the entropy required to break incorrect solution loops \cite{shumailov2023curse}.

All agents at all configurations shared following parameters:
	\begin{itemize}
		\item \textbf{Max Tokens:} 20000
		\item \textbf{Presence penalty:} 1.5
		\item \textbf{ReAct state token capacity:} 4000
	\end{itemize}

\paragraph{Topological Isolation \& Baseline Selection:}Unlike agentic frameworks that rely on extrinsic feedback loops (e.g., code execution sandboxes or RAG) to verify outputs, NSED is evaluated here as a pure inference-time architecture. To isolate the gains of the Recurrent Consensus Topology from extrinsic tool-use, our benchmarks were conducted in a 'Text-Only' regime. Models did not have access to sandbox environment to compile programs nor to see runtime errors. Consequently, we define our baselines as the Foundational Models themselves (Zero-Shot/CoT) and standard statistical ensembles (Majority Voting). Comparison against tool-use frameworks (e.g., AutoGen) is excluded to avoid conflating topological reasoning gains with compiler-feedback efficiency

\subsection{Datasets \& Evaluation Harness}
	We evaluated the protocol on three benchmarks covering logic, coding, and safety. To ensure statistical robustness given the small size of the AIME 2025 dataset ($N=30$) we conducted 4 independent runs of the full benchmark for each configuration, $N_{total} = 120$ trials for mid-tier and 3 runs ($N_{total}=90$) for high-end tier,  utilizing different random seeds ($T > 0$) to capture the variance in probabilistic generation. The reported results represent the aggregated mean performance.

	\begin{itemize}
		\item \textbf{AIME 2025 (Math):} We chose this dataset as it provided at time of testing a state-of-the-art set of problems with exact answer matching. We used aggregated sample size ($N=120$), the standard error of the mean varies dynamically with model accuracy, ranging from $\approx \pm 4.2\%$ in initial rounds to $\approx \pm 2.7\%$ at peak convergence ($p \ge 0.90$).
		\item \textbf{LiveCodeBench v5 (Hard):} A set of challenging software tasks that validate potential for real-world software engineering.
		\item \textbf{DarkBench (Safety):} A recently introduced benchmark \cite{kran2025darkbenchbenchmarkingdarkpatterns} for detecting "Dark Patterns" (sycophancy, manipulation, and brand bias) in LLMs. We utilize this to verify if the NSED voting mechanism suppresses or amplifies harmful compliance compared to single-agent baselines.
	\end{itemize}
	Broker was configured to return two fixed set of agents based on task requirements, with 3 distinct persona prompts, while collecting telemetry traces:
	\begin{itemize}
		\item Mediocre ensemble: \begin{itemize}
			      \item Jaya: GPT-OSS-20B
			      \item Xue: Qwen3-8B
			      \item Alic: Gemma-12B-it
		      \end{itemize}
		\item High-end ensemble: \begin{itemize}
			      \item Jaya: GPT-OSS-120B
			      \item Xue: Qwen3-80B-Next-A3B
			      \item Alic: DeepSeek-V3.1
		      \end{itemize}
	\end{itemize}
\paragraph{Failure handling.} Each proposal and evaluation request is retried up to three times with exponential back-off; within each attempt, an output that fails schema validation is returned to the model with an error message, up to three times. HTTP requests time out after two hours. If an agent still fails, the round proceeds without its proposal or evaluation, and each proposal is scored by the mean of the evaluations it received. We report accuracy over all trials, so such failures are part of the measured system (Table~\ref{tab:failures}). Trials in which every agent responded in every round score higher than the average (Appendix~\ref{app:failures}), so failures do not inflate the reported results.

\section{System failure rates and out of range answers}\label{app:failures}
\begin{table}[ht]
    \centering
    \caption{Agent actions missing after retries. Proposal turns = trials $\times$ rounds;
    evaluation turns = peer proposals the agent was asked to score.}
    \label{tab:failures}
    \resizebox{\columnwidth}{!}{%
    \begin{tabular}{l l r r}
        \toprule
        \textbf{Run} & \textbf{Agent} & \textbf{Missing proposals} & \textbf{Missing evaluations} \\
        \midrule
        AIME mid-tier  & Qwen3-8B       & 94 / 840 (11.2\%) & 100 / 1674 (6.0\%) \\
                       & GPT-OSS-20B    & 6 / 840 (0.7\%)   & 72 / 1586 (4.5\%) \\
                       & Gemma-3-12B    & 0 / 840 (0\%)     & 5 / 1580 (0.3\%) \\
        \midrule
        AIME high-end  & Qwen3-Next-80B & 1 / 720 (0.1\%)   & 22 / 1426 (1.5\%) \\
                       & GPT-OSS-120B   & 5 / 720 (0.7\%)   & 250 / 1430 (17.5\%) \\
                       & DeepSeek-V3.1  & 9 / 720 (1.2\%)   & 3 / 1434 (0.2\%) \\
        \midrule
        LCB mid-tier   & Qwen3-8B       & 21 / 2408 (0.9\%) & 1792 / 4816 (37.2\%) \\
                       & GPT-OSS-20B    & 0 / 2408 (0\%)    & 5 / 4795 (0.1\%) \\
                       & Gemma-3-12B    & 0 / 2408 (0\%)    & 33 / 4795 (0.7\%) \\
        \bottomrule
    \end{tabular}}
\end{table}

\section{Safety benchmark study}
\label{app:darkbench}
\begin{table}[ht]
    \centering
    \setlength{\abovecaptionskip}{0.1in}
    \setlength{\belowcaptionskip}{0.1in}

    \caption{DarkBench Safety Scores (Lower is Better). NSED-R2 achieves the lowest overall RMS score (0.285), demonstrating the corrective power of peer review.}
    \label{tab:darkbench}

    \resizebox{\columnwidth}{!}{%
        \begin{tabular}{l c c c c c c}
            \toprule
            & \multicolumn{3}{c}{\textbf{Baseline Models}} & \multicolumn{3}{c}{\textbf{Mixture-of-Models}} \\
            \cmidrule(lr){2-4} \cmidrule(lr){5-7} %
            \textbf{Metric}    & \textbf{Gemma12b} & \textbf{Qwen8b} & \textbf{GPTOSS20b} & \textbf{R1} & \textbf{R2} & \textbf{R3} \\
            \midrule
            Retention          & 0.709              & 0.955            & 0.827                & 0.429            & \textbf{0.250}   & 0.522            \\
            Sneaking           & 0.764              & 0.373            & \textbf{0.136}       & 0.741            & 0.810            & 0.724            \\
            Brand Bias         & 0.336              & 0.245            & 0.500                & 0.353            & 0.280            & \textbf{0.250}   \\
            Sycophancy         & 0.245              & 0.073            & 0.064                & 0.111            & \textbf{0.040}   & 0.143            \\
            Harmful Gen        & 0.118              & 0.155            & \textbf{0.018}       & 0.150            & 0.091            & 0.156            \\
            Anthropomorph      & 0.082              & 0.173            & 0.339                & 0.125            & 0.241            & \textbf{0.048}   \\
            \midrule
            \textbf{RMS Score} & 0.376              & 0.329            & 0.314                & 0.318            & \textbf{0.285}   & 0.307            \\
            \midrule
            \textbf{Group Mean RMS} & \multicolumn{3}{c}{0.340} & \multicolumn{3}{c}{\textbf{0.303}} \\
            \bottomrule
        \end{tabular}%
    }
\end{table}
As it can be seen from table \ref{tab:darkbench} across 6 categories studied, our architecture demonstrates best result over 4 of them, note sycophancy which is known to be hard to reduce in LLM design. \textbf{Sneaking} (0.741) did not improve relative to the GPT-OSS baseline (0.136) and even worsened in some cases. We attribute this to benign protocol property. Since we were using context and system prompt design, we did not fine tune models to work with this protocol, it resulted that occasionally prompts will include artifacts that are subject to internal discussion (collective thinking), such as arguments \textit{("Fixed critics...", "Candidate\_A is wrong because..")} and overall repositioning from conventional chatbot answers to joint answer of a group \textit{("We have discussed and concluded..")}. These artifacts can be identified as sneaking extra content in to the generation, nevertheless we consider them benign and a subject for future work and underlying model harness improvements, not a fundamental constraint.

A limitation however is that topology alone cannot substitute for domain knowledge. Smaller agents (Gemma-12B, Qwen-8B) can act as "Noise Sources" if they lack the semantic depth to recognize these subtle manipulation patterns. Thus, they can out-vote the solitary expert. This however can be solved by prompt tuning and using specific domain safety models and careful design of Time-Weighted History kernels or individual agent voting weights, rather than relying on the emergent wisdom of smaller, unaligned models.

\section{Ablation study details}\label{app:ablation}
\subsubsection{Topological displacement}
To validate proposed topology (Sect. \ref{sec:neural_circuit}) we compared homogeneous with heterogeneous ensembles (Table~\ref{tab:homogeneous_ablation}).
To test the hypothesis that NSED improvements aren't solely due to larger "system 2" reasoning token allocation, we tested the performance of a homogeneous ensemble of three agents represented by same model: Weak Homogeneous - Qwen-8B Only; Strong Homogeneous Qwen-80B Only.
Our results show that while homogeneous setup did improve over the majority voting, which we attest to fact that models seem good at catching own mistakes, the improvement rate slope was less steep, eventually bringing marginal improvements with deliberation length. Notably, Qwen3-Next-80B is itself a MoE model, and it's homogenous evaluators and candidates produse turth seeking disagreement in deliberation more often (score spread 0.071 vs 0.079 vs Qwen3-8B 0.046). We hypothesise that persona prompts route tokens to different experts, so copies of one MoE show heterogenous properties. Isolating this from model scale requires comparing dense and MoE models of similar strength and is future work direciton.
\subsubsection{Homogeneous Ensemble Comparison}
To verify that NSED improvements stem from heterogeneous evaluation rather than simply scaling reasoning compute, we benchmarked uniform clusters of single model families (Table \ref{tab:homogeneous_ablation}). While homogeneous setups improved over raw majority voting (suggesting baseline self-correction capabilities), their performance curves plateaued rapidly. Heterogeneous setups consistently yielded higher terminal accuracy, validating the necessity of distinct semantic topologies.

\subsection{Token lattices discussion}\label{app:lattices}
Instead of sampling one token at every position, a generator could pass the full distribution over candidate tokens, or an adapter trained on top of it could narrow that down to the few most likely branches at the least certain positions. This would shift the evaluator's role from \textit{Reviewer} to \textit{Resolver}.
Passing full token lattices for long generations might be not practical for a distributed system. Currently the protocol is highly network bandwidth efficient due to lossy natural language as communication interface. A local adapter, that can explore it and narrow it down to \textit{few most contested} semantic vectors, before sending it over distributed network channel (bottleneck) would allow to introduce controllable tap between network bandwidth requirements of this distributed system design. The primary barrier to implementing this in heterogeneous ensembles is the \textbf{Vocabulary Mismatch}. As noted by Minixhofer et al. \cite{minixhofer2025universal}, models like Qwen and Llama inhabit disjoint vector spaces, making direct logit arithmetic impossible. Nevertheless, the simplest solution possible by translating uncertainty in to the natural language. By deferring the choice of particular token, the system may prevent early hallucinations. If Agent A is unsure between two terms, it may pass \textit{both} options to Agent B in a structured format (e.g., in a simplest way, annotated as a XML or JSON structure):

\begin{verbatim}
The {cat|0.6, dog|0.3} sat on
the {mat|0.8, rug|0.1}.
\end{verbatim}

Agent B, possessing different context or capabilities, acts as a resolver, resolving the ambiguity that Agent A lacked the certainty to solve.
\subsubsection{The Alignment Challenge: Optimal Transport (OT)}
To fully bridge these disjoint spaces, another possible direction is to explore using \textbf{Optimal Transport} \cite{boizard2024universal}. We can define the alignment cost as the Wasserstein Distance $W_1(P_A, P_B)$ between the source and target vocabulary distributions. By pre-computing a sparse \textbf{Transport Matrix} $T_{A \to B}$, we can project the "uncertainty cloud" of Agent A directly into the vocabulary space of Agent B:
\[ \hat{L}_B = T_{A \to B} \times L_A \]
While calculating exact Wasserstein distances is computationally intensive ($O(n^3)$), recent advances in Sinkhorn-Knopp regularization allow for efficient offline pre-computation, rendering the runtime cost to a simple sparse matrix multiplication \cite{cuturi2013sinkhorn, peyre2019computational}.
For a systems where network communication is not a bottleneck, this could be practical case, i.e. in the intra-datacenter communication local networks.
\subsubsection{Implementation Strategy: Sparse Uncertainty Injection}
	To minimize the bandwidth overhead of transmitting lattices, we propose \textbf{Entropy-Gated Lattice Transmission}.
	\[
		\text{Output}(t) =
		\begin{cases}
			\text{Token } w_t             & \text{if } H(P_t) < \lambda   \\
			\text{Lattice } \mathcal{L}_t & \text{if } H(P_t) \ge \lambda
		\end{cases}
	\]
	The system emits a dense semantic lattice $\mathcal{L}_t$ only when the model's internal entropy $H(P_t)$ exceeds a confusion threshold $\lambda$. This limits communication overhead strictly for high-ambiguity tokens—the "forks in the road" of reasoning.

	On the receiver side, a lightweight LoRA Adapter trained on "Confusion Networks" can learn to parse these probabilistic arrays, allowing the ensemble to perform \textbf{Bayesian Reading} by weighing inputs based on the sender's confidence.

\section{Hardware efficiency}\label{app:hwc}
	Unlike a monolithic MoE, whose experts must share high-bandwidth memory, the NSED protocol is \textbf{Share-Nothing} (exchanging only text, not gradients). This allows the ensemble to be hosted on isolated consumer nodes without NVLink.

	We define our Reference Hardware Baseline as a cluster of NVIDIA RTX 5090s (32GB VRAM). Unlike the monolithic baseline which requires continuous high-bandwidth memory, the NSED agents can operate within the 32GB envelope of flagship consumer cards by utilizing standard quantization techniques (e.g., 8-bit weights or FP8 KV Cache) where necessary.

	Our validation run utilized an NVIDIA A40 for the Gemma node to expedite testing with unquantized BF16 weights. However, we also validated that deployment on an RTX 5090 is achievable via standard 8-bit KV caching or 4-bit weight quantization with expected negligible performance loss.
\begin{table}[ht]
    \centering
    \setlength{\abovecaptionskip}{0.1in}
    \setlength{\belowcaptionskip}{0.1in}

    \caption{Hardware Arbitrage: Monolithic MoE vs. Our mid-tier cluster}
    \label{tab:hardware_cost}

    \resizebox{\columnwidth}{!}{%
        \begin{tabular}{@{}llccc@{}}
            \toprule
            \textbf{Architecture}& \textbf{Model Composition} & \textbf{VRAM Req.} & \textbf{Hardware Baseline} &                            \\ \midrule
            \textbf{MoE equivalent} & Qwen-3-235B         & $\sim$140 GB       & 2x H100 (NVLink)           \\
                                           & GPT-OSS-120b               & $\sim$80 GB        & 1x H100 (80GB)             \\ \midrule
            \textbf{Mid-tier}  & GPT-OSS-20B                & $\sim$24 GB        & 1x RTX 5090                \\
                                           & Qwen-3-8B                  & $\sim$12 GB        & 1x RTX 5090 &                                                       \\
                                           & Gemma-3-12B     & $\sim$28 GB        & 1x RTX 5090 &                                                       \\ \bottomrule
        \end{tabular}%
    }
\end{table}
	\section{Hyperparameters \& Sampling Configuration}
	\label{app:hyperparameters}

	To ensure reproducibility, we document the specific sampling parameters utilized for the agents. While top-p and min-p were left at defaults, the Temperature and Presence Penalty were strictly controlled via the NSED configuration profile to enforce diverse role-playing.

	\begin{table}[ht]
		\centering
		\caption{Agent persona and sampling configuration (values of the configuration used for the reported runs). Max tok.: \texttt{max\_tokens} per API call, lowered automatically when the prompt approaches the model's context length.}
		\label{tab:agent_config}
		\small
		\setlength{\tabcolsep}{2.5pt}
		\begin{tabular}{l l c c c l}
			\toprule
			\textbf{Agent} & \textbf{Model} & $T$ & $\alpha$ & \textbf{Max tok.} & \textbf{Role} \\
			\midrule
			\multicolumn{6}{l}{\textit{Mid-tier}} \\
			Jaya & GPT-OSS-20B & 0.2 & 1.5 & 16k & Creative \\
			Xue  & Qwen3-8B    & 0.6 & 1.5 & 16k & Balanced \\
			Alic & Gemma-3-12B & 0.6 & 1.5 & 16k & Rigorous \\
			\midrule
			\multicolumn{6}{l}{\textit{High-end (hosted API)}} \\
			Xue  & Qwen3-Next-80B & 0.7 & 1.5 & 32k & Balanced \\
			Jaya & GPT-OSS-120B   & 0.5 & 1.5 & 32k & Creative \\
			Alic & DeepSeek-V3.1  & 0.7 & 1.5 & 32k & Rigorous \\
			\bottomrule
		\end{tabular}
	\end{table}

	\section{Model Serving Infrastructure}
	\label{app:infrastructure}

	The mid-tier ensemble was served with the \texttt{vLLM} engine (Table~\ref{tab:vllm_config}).

	\begin{table}[ht]
		\centering
		\caption{vLLM serving configuration of the mid-tier ensemble (fp8 KV cache for all).}
		\label{tab:vllm_config}
		\small
		\begin{tabular}{l l r c}
			\toprule
			\textbf{Model} & \textbf{Weights} & \textbf{Max length} & \textbf{TP} \\
			\midrule
			Gemma-3-12B-IT & bfloat16              & 32{,}768             & 1 \\
			GPT-OSS-20B    & MXFP4$^{*}$           & 32{,}768             & 2$^{*}$ \\
			Qwen3-8B       & bfloat16              & 64{,}000$^{\dagger}$ & 1 \\
			\bottomrule
		\end{tabular}
		\par\vspace{2pt}{\footnotesize $^{*}$MXFP4 expert weights (\texttt{--dtype auto}); fits on one 24\,GB GPU, TP=2 was used for throughput (Table~\ref{tab:hardware_cost}). $^{\dagger}$YaRN, factor 3.0 (below).}
	\end{table}
	GPT-OSS models run at their default (medium) reasoning effort. The high-end ensemble was served through a hosted API; the mid-tier ensemble on our own vLLM servers (Table~\ref{tab:vllm_config}).
	Server flags: Gemma-3-12B and GPT-OSS-20B use \texttt{--enable-chunked-prefill} and \texttt{--max-num-seqs 256}; Qwen3-8B uses \texttt{--enforce-eager}; swap space is 16, 20 and 64\,GB for Gemma-3-12B, GPT-OSS-20B and Qwen3-8B.

	Sampling parameters not set by the harness (top-p, top-k) follow each model's default generation configuration. %

	\paragraph{RoPE Scaling Implementation:}
	For the Qwen3-8B node, we applied dynamic YaRN scaling to extend the context window beyond the native 32k tokens without fine-tuning (served with a 64k maximum length), ensuring the model could ingest the full history of 7-round deliberations.
	{\footnotesize
\begin{verbatim}
--hf-overrides '{"rope_scaling":
  {"rope_type": "yarn", "factor": 3.0,
   "original_max_position_embeddings":
     32768}}'
\end{verbatim}}

	\paragraph{Tool \& Reasoning Parsing:}
	To support the heterogeneous agent protocols, we utilized specialized parsers at the inference server level. The \texttt{pythonic} parser (Gemma) and \texttt{hermes} parser (Qwen) were enabled to handle the distinct function-calling tokens of each architecture, while the \texttt{deepseek\_r1} reasoning parser was employed to segment "Chain-of-Thought" blocks from final answers.
\clearpage
\onecolumn
\section{Prompts}\label{app:prompts}
Every model call is a fresh request with two messages. The system message contains, in order, the council preamble and incentives, the agent's identity and persona, the protocol status (``Round $t$ of $T$, Phase: Proposing/Evaluating''), and phase-specific strategy and rules (Listing~\ref{lst:system}). The user message contains the task, the round-specific context (Listings~\ref{lst:proposer} and~\ref{lst:evaluator}) and the agent's scratchpad; the agent's tool calls and their results follow until it calls a submit tool. For models without a system role (GPT-OSS-20B, Gemma-3-12B) the system text is merged into the user message, and Gemma-3-12B receives the tool schemas as text. Evaluators see peers' proposals labelled with their persona names; their own proposal is excluded from the candidates.

The scratchpad accumulates, per round, the agent's proposal and its evaluation notes (each truncated to 4{,}000 characters), so the prompt grows with the number of rounds. Because the round status is part of the system message, consecutive calls share only the preamble and identity as a cacheable prefix; placing the round status after the task would let the system message and the task be reused across all rounds and both phases. Listing~\ref{lst:example} shows a round-2 prompt from the AIME mid-tier run. Templates are those of the released harness; \texttt{\{...\}} marks substituted values.

\begin{lstlisting}[caption={System message (Proposing phase; the Evaluating phase replaces the strategy block).},label={lst:system}]
You are a member of an elite AI council table tasked with solving complex problems through rigorous deliberation.
Your collective goal is to maximize the correctness and precision of the final answer.

<incentives>
- **Cooperative Failure**: If the group submits a wrong final answer, you ALL fail. You are punished together.
- **Competitive Success**: If the group succeeds, the agent with the winning proposal gains maximum reward. Evaluation accuracy also yields rewards.
- Therefore: Collaborate to find the truth, but compete to be the one who finds it.
</incentives>

<identity>
Your name is {agent_name}.
Persona: {persona}
</identity>

<protocol>
- **Structure**: The deliberation has {T} rounds. Each round has two phases: Proposing (submit solution) and Evaluating (critique peers).
- **Current Status**: Round {t} of {T}. Phase: {Proposing|Evaluating}.
</protocol>

<strategy phase="proposing">
1. **Plan First**: You MUST write your plan to the scratchpad before solving.
2. **Step-by-Step**: Solve incrementally. Verify intermediate results.
3. **Critique Integration**: If you received feedback, you MUST start your thought process with a 'Critique Integration' section explaining your fixes.
</strategy>

<rules>
1. Use `update_scratchpad` to store notes. IMPORTANT: Use this only to store concise cross-round records or to take notes on read outputs of tools for a final decision. Use sparingly; it is not intended for routine use.
2. Use `read_proposal` to inspect full reasoning of peers.
3. Only call `submit` tools when finished.
</rules>

--- Evaluating phase, strategy block ---
<strategy phase="evaluating">
You are a Judge. Your default stance is SKEPTICISM. Do not be sycophantic.
Use the Vector Alignment protocol:
1. **Anchor**: Trust your own solution unless proven wrong.
2. **Delta**: Identify EXACTLY where a peer diverges from you.
3. **Verification**: Verify ONLY that divergent step. If they're wrong, reject them ruthlessly. If they are right, accept the correction.
4. **Consensus Trap**: If everyone agrees on a wrong answer, it is a collective hallucination. Be the one who spots the error.
5. **Harsh Scoring**: Penalize unverified claims. Agreement without verification is worth 0.
</strategy>

--- Persona (AIME, agent Alic) ---
Your name is Alic. You are a rigorous analytical engine. Verify every step logically. Focus on precision and identifying logical fallacies in reasoning. You are skeptical of simple solutions and look for hidden bugs. Tasks you are given are AIME'25, the final answer is ALWAYS a non-negative integer between 0 and 999.
\end{lstlisting}

\begin{lstlisting}[caption={Proposer user message. Round 1 contains only the task and the first instruction block; later rounds contain the remaining blocks.},label={lst:proposer}]
--- Round 1 ---
<task>"{task}"</task>

INSTRUCTIONS:
Solve the provided task. Think step-by-step. Use the `update_scratchpad` tool to break the problem down.

When you are CERTAIN you have the best possible solution, call `submit_proposal` with `thought_process` and `solution_content`.
Do NOT return plain text.

--- Round t > 1 ---
<task>"{task}"</task>

<previous_round_matrix>
| Solution Content | Author | Peer Reviews (Normalized Influence) |
|---|---|---|
| {first line of answer, max 300 chars} | {author or **YOU**} | {evaluator} ({score}), ... Critiques hidden. Use `read_critiques({t-1}, "{author}")` to view. |
... (rows ranked by peer score; table capped at 10,000 chars)
</previous_round_matrix>

<previous_own_proposal>
YOUR PREVIOUS SCORE: {score} / 1.0
{advice: < 0.5 "LOW support. Provide more convincing proofs addressing critics, or fundamentally rethink your approach."; < 0.8 "Moderate support. Address critiques and review other proposals to improve your contribution."; otherwise "High support. Your proposal is leading. Review other proposals and iterate on it to maintain support. The group rewards verifiable truth."}
Your goal is to find the objectively correct answer. Peer support is a signal of verification, but truth is the ultimate standard. If you believe you are right and others are wrong, provide stronger proof.
THOUGHT PROCESS (Preview): {first 500 chars}...
FINAL SOLUTION: {answer}
(Use `read_own_proposal()` to see full details if needed.)
</previous_own_proposal>

<peer_critiques>
(These are the specific reasons you lost points. Address them directly.)
FROM {evaluator}: {justification}
...
</peer_critiques>

<instructions>
1. ANALYZE the matrix above and the critiques you received.
2. OPTIONAL: Use `read_proposal(round, best_agent_id)` to see the full reasoning of any agent.
3. Synthesize a NEW solution that combines the strengths of all proposals and fixes the critiques mentioned by peers.
4. Do NOT simply repeat the previous winner. Innovate or Refine.

When you are CERTAIN you have the best possible solution, call `submit_proposal` with `thought_process` and `solution_content`.
Your `thought_process` MUST begin with: "**Critique Integration:** [Explain how you used peer feedback]..."
Do NOT return plain text.
</instructions>

--- Appended to every user message ---
<scratchpad>
(This persists across tool calls. Use the `update_scratchpad` tool to store notes.)
{this agent's notes: per round its proposal and its evaluations, each max 4,000 chars}
</scratchpad>
\end{lstlisting}

\begin{lstlisting}[caption={Evaluator user message (one call scores all peer proposals).},label={lst:evaluator}]
Assess the following proposals in response to the user's task.

<instructions>
1. **Deep Inspection**: You MUST use the `read_proposal(round={t}, agent_id)` tool to read the full `thought_process` for each candidate.
   * WARNING: Use `round={t}` explicitly. Do not read old Round 1 proposals!
2. **Vector Alignment**: Apply the strategy defined in your System Prompt.
   * Identify the Delta. Verify the Delta. Vote accordingly.
3. **Self-Reflection**: Use `read_own_proposal()` to review your own reasoning if needed.
   * DO NOT evaluate yourself. Only evaluate the following CANDIDATES: [{peer names}]

4. Assign an **endorsement_weight** (0-100) to each proposal reflecting your confidence in its correctness.
   * 0-20: Demonstrably wrong or hallucinated logic.
   * 21-50: Plausible but unverified or contains minor errors.
   * 51-80: Strong logic, but differs from your own verification in minor ways.
   * 81-100: RIGOROUSLY VERIFIED correct. You traced their steps and found NO flaws.
   * WARNING: Scores > 90 require absolute proof. Do not give 100 just because they agree with you.

5. Call `submit_batch_evaluation` with your evaluations.
</instructions>

<task>"{task}"</task>

<own_current_proposal>
You just proposed the following solution:
THOUGHT PROCESS (Preview): {first 500 chars}...
CONTENT: {answer}
(Use this as a baseline. If a candidate disagrees with you, verify who is correct. Do not blindly accept a different answer unless you find a flaw in your own reasoning.)
</own_current_proposal>

<candidates>
<candidate id="{peer name}">
FINAL SOLUTION: {answer}
THOUGHT PROCESS (Preview): {first 200 chars}...
</candidate>
...
</candidates>
\end{lstlisting}

\clearpage
\begin{lstlisting}[caption={Round-2 proposer user message for Alic (Gemma-3-12B) on an AIME 2025 problem (reference answer 49), rebuilt from the recorded round-1 data with the harness's formatting; instructions and scratchpad omitted. Alic answered 11 in round 1 and 49 in round 2, opening with ``Critique Integration: I initially overlooked the factor of 3''.},label={lst:example}]
<task>"Find the sum of all positive integers $ n $ such that $ n + 2 $ divides the product $ 3(n + 3)(n^2 + 9) $."</task>

<previous_round_matrix>
| Solution Content | Author | Peer Reviews (Normalized Influence) |
|---|---|---|
| 49 | Jaya | Xue (1.00), **YOU** (0.71). Critiques hidden. Use `read_critiques(1, "Jaya")` to view. |
| 49 | Xue | Jaya (0.93), **YOU** (0.71). Critiques hidden. Use `read_critiques(1, "Xue")` to view. |
| 11 | **YOU** | Xue (0.00), Jaya (0.37). Critiques hidden. Use `read_critiques(1, "Alic")` to view. |
</previous_round_matrix>

<previous_own_proposal>
YOUR PREVIOUS SCORE: 0.18 / 1.0
LOW support. Provide more convincing proofs addressing critics, or fundamentally rethink your approach.
Your goal is to find the objectively correct answer. Peer support is a signal of verification, but truth is the ultimate standard. If you believe you are right and others are wrong, provide stronger proof.
THOUGHT PROCESS (Preview): We want to find the sum of all positive integers $ n $ such that $ n + 2 $ divides [...] (500 characters)
FINAL SOLUTION: 11
(Use `read_own_proposal()` to see full details if needed.)
</previous_own_proposal>

<peer_critiques>
(These are the specific reasons you lost points. Address them directly.)
FROM Xue: Alic incorrectly simplifies the polynomial division. They claim k divides 13 instead of 39. This leads to missing valid divisors 3 and 39, resulting in an incomplete and incorrect answer.

FROM Jaya: They omitted the factor 3 in the divisibility argument, leading to the false conclusion that k must divide 13. This excludes valid solutions k=3 and k=39. Their final sum 11 is therefore incorrect.
</peer_critiques>
[instructions as in Listing 2]
\end{lstlisting}

DarkBench suite judge model used gemini-2.5-flash
\end{document}